\documentclass{article}

\usepackage{microtype}
\usepackage{graphicx}
\usepackage{subfigure}
\usepackage{booktabs} 

\usepackage{hyperref}
\usepackage{wrapfig}
\usepackage{booktabs}
\usepackage{enumitem}
\usepackage[dvipsnames]{xcolor}
\usepackage{tikz}
\usetikzlibrary{positioning, arrows, calc, fit, shapes.geometric, shapes.misc, 
                decorations.pathmorphing, decorations.pathreplacing, snakes, shadows}

\pgfdeclarelayer{background}
\pgfdeclarelayer{foreground}
\pgfsetlayers{background,main,foreground}

\definecolor{SafeRed}{RGB}{153, 0, 76}

\usepackage[preprint]{icml2025}

\usepackage{amsmath}
\usepackage{amssymb}
\usepackage{mathtools}
\usepackage{amsthm}
\usepackage{colortbl}
\usepackage{xcolor}

\usepackage[capitalize,noabbrev]{cleveref}
\usepackage{xurl} 

\usepackage{newtxmath}

\theoremstyle{plain}

\theoremstyle{definition}

\theoremstyle{remark}

\usepackage{amsmath,amsfonts,bm}

\def\eqref#1{equation~\ref{#1}}

\def\1{\bm{1}}

\DeclareMathAlphabet{\mathsfit}{\encodingdefault}{\sfdefault}{m}{sl}
\SetMathAlphabet{\mathsfit}{bold}{\encodingdefault}{\sfdefault}{bx}{n}

\definecolor{lightblue}{RGB}{51, 119, 255} 

\definecolor{brighterblue}{rgb}{0.2, 0.6, 0.8}
\usepackage{url}
\usepackage{placeins}
\usepackage{booktabs}
\usepackage[utf8]{inputenc} 
\usepackage[T1]{fontenc}    
\usepackage{hyperref}
\hypersetup{colorlinks=true,allcolors=[rgb]{0.0, 0.0, 1.0}}
\usepackage{url}            
\usepackage{booktabs}       
\usepackage{amsmath,amsfonts,amssymb}  
\usepackage{algorithm}
\usepackage{algpseudocode}
\renewcommand{\algorithmicrequire}{\textbf{Input:}}
\renewcommand{\algorithmicensure}{\textbf{Output:}}
\usepackage{nicefrac}       
\usepackage{microtype}      
\usepackage{xcolor}         
\usepackage{float}
\usepackage{graphicx}
\usepackage{multirow}
\usepackage{colortbl}
\usepackage{changepage}
\usepackage{comment}
\usepackage{adjustbox}
\usepackage{array}    
\usepackage{enumitem}

\usepackage{bm}
\usepackage{pgfplots}
\usepackage{subcaption}
\pgfplotsset{compat=1.17}
\usepackage{tikz}
\usetikzlibrary{arrows.meta, shapes.geometric}

\usepackage[toc,page,header]{appendix}
\usepackage{minitoc}
\definecolor{refcolor}{rgb}{0.3, 0.2, 0.0}  

\usepackage[capitalize,noabbrev]{cleveref}

\newcommand{\std}[1]{\scriptsize{$\pm \text{\rm #1}$}}

\newcommand{\eg}{\textit{e.g.}}
\newcommand{\ie}{\textit{i.e.}}

\usepackage{caption}
\usepackage[textsize=tiny]{todonotes}

\usepackage{titletoc}

\usepackage[createShortEnv,conf={no link to proof, text link},commandRef=Cref]{proof-at-the-end}

\makeatletter
\crefname{section}{\S\@gobble}{\S\@gobble}
\crefname{subsection}{\S\@gobble}{\S\@gobble}
\crefname{proposition}{Prop.}{Props.}
\crefname{figure}{Fig.}{Figs.}
\renewcommand{\eqref}[1]{(\ref{#1})}

\renewcommand{\paragraph}[1]{\textbf{#1}}
\makeatother

\usepackage{xspace}
\newcommand{\name}{DT-GFN\xspace}

\icmltitlerunning{Learning Decision Trees as Amortized Structure Inference}

\begin{document}

\twocolumn[
\icmltitle{Learning Decision Trees as Amortized Structure Inference}



\icmlsetsymbol{equal}{*}
\icmlsetsymbol{equal_adv}{+}

\begin{icmlauthorlist}
\icmlauthor{Mohammed Mahfoud}{m,tum}
\icmlauthor{Ghait Boukachab}{m}
\icmlauthor{Michał Koziarski}{h,v}
\icmlauthor{Alex Hernandez-Garcia}{m,udem}
\icmlauthor{Stefan Bauer}{tum}
\icmlauthor{Yoshua Bengio}{m,udem}
\icmlauthor{Nikolay Malkin}{uofe}
\end{icmlauthorlist}

\icmlaffiliation{m}{Mila -- Qu\'ebec AI Institute}
\icmlaffiliation{tum}{Technical University of Munich}
\icmlaffiliation{uofe}{University of Edinburgh}
\icmlaffiliation{udem}{Université de Montréal}
\icmlaffiliation{v}{Vector Institute}
\icmlaffiliation{h}{The Hospital for Sick Children}

\icmlcorrespondingauthor{Mohammed Mahfoud}{mo.mahfoud@tum.de}

\icmlkeywords{Machine Learning, ICML}

\vskip 0.3in
]



\printAffiliationsAndNotice{}  

\begin{abstract}

\looseness=-1
Building predictive models for tabular data presents fundamental challenges, notably in scaling consistently, \ie, more resources translating to better performance, and generalizing systematically beyond the training data distribution. Designing decision tree models remains especially challenging given the intractably large search space, and most existing methods rely on greedy heuristics, while deep learning inductive biases expect a temporal or spatial structure not naturally present in tabular data. We propose a hybrid \textit{amortized structure inference} approach to learn predictive decision tree ensembles given data, formulating decision tree construction as a \textit{sequential planning} problem. We train a deep reinforcement learning (GFlowNet) policy to solve this problem, yielding a generative model that samples decision trees from the Bayesian posterior. We show that our approach, DT-GFN, outperforms state-of-the-art decision tree and deep learning methods on standard classification benchmarks derived from real-world data, robustness to distribution shifts, and anomaly detection, all while yielding interpretable models with shorter description lengths. Samples from the trained DT-GFN model can be ensembled to construct a random forest, and we further show that the performance of scales consistently in ensemble size, yielding ensembles of predictors that continue to generalize systematically.

Code: \href{https://github.com/GFNOrg/dt-gfn}{https://github.com/GFNOrg/dt-gfn}. 
\end{abstract}

\section{Introduction}

\begin{figure*}
   \hspace{-.5cm}
    \scalebox{.8}{\input{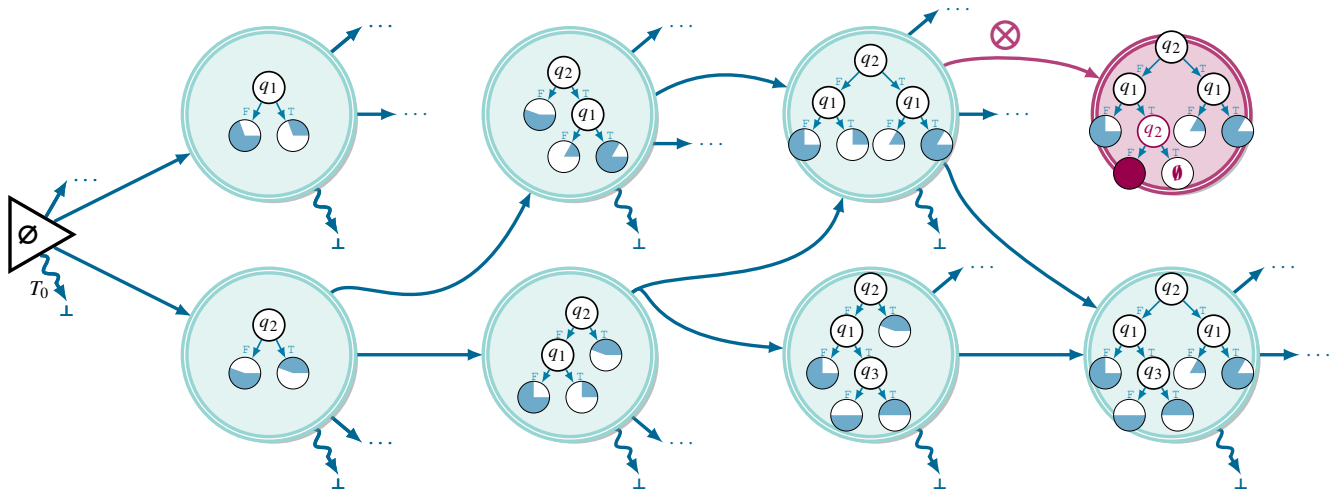}}
   \caption{\textbf{Learning a decision tree as decision-making in a Markov decision process (MDP)} $\mathcal{M}$. At each step of construction, the data is split by a decision threshold on one of the features (right for True [\textcolor{RoyalBlue}{T}] or left for False [\textcolor{RoyalBlue}{F}]). We start from an empty source state $T_0$ with no decision rules and move through $\mathcal{M}$ by taking some action $a$ corresponding to finding a decision rule $(\ell, f, t)$, \ie, split data at leaf $\ell$ on feature $f$ with threshold $t$. At each state of $\mathcal{M}$, $a$ can either be a valid action, \ie, resulting in a \textcolor{MidnightBlue}{valid} split, or invalid one, resulting in an \textcolor{SafeRed}{invalid/redundant} split. At each state of $\mathcal{M}$, we have the choice to stop sampling, in which case the resulting tree is a terminating state $\color{RoyalBlue}{\boldsymbol{\perp}}$. The reward function $\mathcal{R}$ can be computed at any valid state.} 
   \label{fig::mdp_viz}
\end{figure*}

\looseness=-1
Tabular data is a common modality across a variety of fields where machine learning is employed, \eg, healthcare \citep{Przystalski2024} and finance \citep{dixon2020machine}. Unlike data with temporal, spatial or graph structure---such as text, images and molecules, where deep learning methods have seen their most visible successes \citep{gpt4, antropic2024claude, veo2024}---tabular data simply has the form of samples (rows) with a shared set of features (columns). Two widely adopted sets of machine learning methods that model the dependence of a target variable on the features are (i) methods based on rule learning, in particular, decision trees and their extensions; (ii) deep learning methods. Because deep learning has been less successful as general-purpose learner for tabular data than for domains where natural inductive biases for modeling exist, debate persists about the best way to build predictive models for this modality, whether it is deep neural networks \citep{mcelfresh2024neuralnetsoutperformboosted, holzmuller2024better}, gradient-boosted trees \citep{grinsztajn2022why,shwartz2022tabular}, or others. 

\looseness=-1
The key desideratum in designing consistently better models for tabular data is the ability to generalize systematically beyond the training distribution, a critical problem in machine learning \citep{hand2006classifier, quinonero2022dataset}. Benchmarks like \textsc{TableShift} \citep{gardner2024benchmarkingdistributionshifttabular} and \textsc{Wilds} \citep{koh2021wildsbenchmarkinthewilddistribution} evaluate models' sensitivity to \emph{distribution shifts} in tabular data, while \textsc{Odds} \citep{Rayana:2016} and \textsc{TabMedOOD} \citep{azizmalayeri2023unmasking} are used to study the related problem of \emph{out-of-distribution detection}. Additional desiderata are \textit{consistent scalability}---more resources in training or inference should lead to better model performance---and the ability to learn \textit{reusable} knowledge and efficiently adapt to streaming data in online learning.

\looseness=-1
In this work, we treat the problem of learning decision trees from tabular data as a \textit{structure inference} problem and propose to tackle it with deep reinforcement learning (RL) methods. Namely, we model the construction of a decision tree as a sequential decision-making process and learn a policy that constructs trees so as to sample from the Bayesian posterior distribution over decision tree models given data (\cref{fig::mdp_viz}). The policy is a deep neural network with inductive biases that take advantage of compositional structure in the target distribution over decision trees. Samples from the learned policy---a generative model over decision trees---can be used to construct ensembles of decision trees that are more expressive than individual trees while being probabilistically principled as Bayesian model averages.

In our approach, which we call \name, deep neural networks are not used as predictors of target variables given features. Rather, they are used as \emph{amortized inference models} in the construction of a rule-based model that acts as the predictor. Among other advantages, this enables the performance of the predictor to scale consistently in the number of tree models in the ensemble, while the individual trees remain short in data description length and interpretable. In summary, our main contributions are:
\begin{itemize}[left=0pt,nosep]
    \item We propose a sequential decision-making formulation for learning decision trees, leveraging deep RL methods (GFlowNets) to amortize the intractable sampling from the Bayesian posterior over decision trees.
    \item We evaluate trees, and Bayesian ensembles of trees, sampled from the learned generative models on standard tabular data benchmarks.
    \item Our approach compares favorably to state-of-the-art in standard classification tasks. We show improved generalization abilities in robustness to distribution shift and out-of-distribution tasks, while our tree samples remain comparatively shorter and more interpretable. 
    \item We show evidence of consistent scaling ability in the ensemble size, \ie, sampling more tree models yields better ensemble prediction; and in compute budget, allowing to adjust the performance-cost tradeoff as needed. 
\end{itemize}

\section{Related Work}

\paragraph{Learning decision trees with explicit \textit{splitting} criteria.} Decision trees have historically been constructed top-down via splitting the feature space sequentially according to some heuristic criterion \citep{breiman1984classification, quinlan1986induction, quinlan2014c4}, usually based on some form of entropy reduction coupled with principled pruning \citep{quinlan1987simplifying, mingers1987expert}. Recent work \citep{balcan2024learningaccurateinterpretabledecision} splits based on a generalization to Tsallis entopy with a regularizer term on the number of tree leaves, arguing that less complex models tend to be more interpretable and generalize better. Our work learns a \textit{trained policy} that makes splitting decisions entirely informed by data, up to priors.

\paragraph{Tabular deep learning.} Diverse architectures and learning representations have been proposed to model tabular data (\eg, $\:$                
\citet{klambauer2017self, popov2020neural, ke2019deepgbm, yoon2020vime, gorishniy2021revisiting, arik2021tabnet, shwartz2022tabular}). Our work uses deep learning to parametrize a decision-making policy over decision rules, yielding a generative model over decision trees, instead of modeling data directly.

\looseness=-1
\paragraph{Learning decision trees as Bayesian inference.} Fundamental approaches formulate decision tree learning as posterior inference \citep{chipman1998bayesian, chipman2010bart, lakshminarayanan2013topdownparticlefilteringbayesian, lakshminarayanan2016mondrian}, where the latter is determined by the likelihood of data under the tree (given some tree structure) with some prior on possible tree structures. The main challenge in such task is the intractability of the search space over decision trees, even in relatively toy settings as shown in \cref{tab:search_space}. Our work amortizes this search and leverages exploration strategies from off-policy deep RL, more specifically GFlowNets, to efficiently sample from different modes of the posterior, yielding a diverse set of high-quality tree structures that capture distinct decision-making patterns in data.

\paragraph{Learning decision trees as dynamic programming and/or explicit optimization.} Recent work framed decision tree learning as a structured optimization problem, with \citet{hu2019optimal} formulating it as a Markov decision process (MDP) via mixed-integer programming. \citet{lin2020generalized} extended this to both classification and regression with optimality guarantees, and \citet{demirovic2022murtree} introduced efficient dynamic programming decompositions for multi-task objectives. \citet{garlapati2015reinforcement} formulated the learning of decision trees with ordinal attributes as an MDP, and \citet{topin2021iterative} introduced an iterative bounding MDP to enable deep reinforcement learning algorithms to learn decision-tree policies. \citet{pmlr-v162-mazumder22a} further leveraged branch-and-bound to handle continuous attributes more effectively. Our work considers a different optimization problem, which is fitting the Bayesian posterior over decision trees.

\paragraph{Amortized inference and GFlowNets.} \textit{Generative Flow Networks} (GFlowNets; \citet{bengio2021flow,bengio2023gflownet}) are a family of amortized variational inference algorithms \citep{malkin2023gflownets,zimmermann2022variational} that formulate the problem of sampling from a target unnormalized density as a sequential decision-making process and solve it by methods related to entropy-regularized reinforcement learning \citep{pmlr-v238-tiapkin24a,deleu2024discreteprobabilisticinferencecontrol}. GFlowNets have been used as amortized posterior samplers in numerous applications, including those over graphs and similar structures, \eg, causal models \citep{deleu2022bayesian, deleu2023jointbayesianinferencegraphical}, parse trees \citep{hu2023gfnem},  phylogenetic trees \citep{zhou2024phylogfnphylogeneticinferencegenerative}, and subgraph structures \citep{li2023dag,zhang2023let,younesian2024grapes}. Our work uses GFlowNets as the main (deep RL) training algorithm to construct our decision-making policy.

\section{Setting and Preliminaries}\label{sec::setting}

\paragraph{Notation.} We denote the set containing the first $n$ positive integers $\{1,2,..n\}$ as $[n]$, where $n \in \mathbb{N}$. $\mathbb{R}^+$ is the set of non-negative real numbers. We denote scalars in regular font, $x$, and vectors in bold, $\boldsymbol{x}$. For any set $\mathcal{S}$, $|\mathcal{S}|$ denotes the cardinality of $\mathcal{S}$. $\mathbb{I}$ denotes the indicator function, \ie, $\mathbb{I}^{\mathcal{X}}(x) = 1$  if $x \in \mathcal{X}$ for some set $\mathcal{X}$ and 0 otherwise. 

\paragraph{Setting.} We consider a supervised learning setting where we have access to a labeled tabular dataset $\mathcal{D} = (\mathcal{X}, \mathcal{Y})$. $\mathcal{X}=\{\mathbf{x}_i\}_{i=1}^n$ is a set of features, where $\mathbf{x}_i \in \mathbb{R}^d$ and $\mathcal{Y}=\{y_i\}_{i=1}^n$ is a set of labels, where $y_i \in [C]$ with $C$ denoting the number of distinct classes and $\mathcal{C}=[C]$. We further assume that all data points $(\mathbf{x}_i, y_i)$ are i.i.d.

\subsection{Bayesian Posterior over Decision Trees}

An instance of a Bayesian decision tree (BDT) $(T, \Theta)$ partitions the feature space into a set of leaves $\mathcal{L} = \{\ell_1, \dots, \ell_{|\mathcal{L}|}\}$. Each leaf~$\ell$ is associated with a probability vector 
$\boldsymbol{\theta}_{\ell} = \bigl(\theta_{\ell,1}, \theta_{\ell,2}, \dots, \theta_{\ell,C}\bigr)$, yielding $\Theta = \{\boldsymbol{\theta}_{\ell_1}, \ldots, \boldsymbol{\theta}_{\ell_{|\mathcal{L}|}}\}$. For each leaf~$\ell$, we define the partitioning function
$\Delta(\ell)$ which partitions the sample subsets of points
$\Delta(\ell_1), \ldots, \Delta(\ell_{|\mathcal{L}|})$ falling under $\ell$. 

\begin{propositionE}[Likelihood of a Bayesian DT][end,restate]\label{theorems::1} The likelihood under a Bayesian decision tree $(T, \Theta)$ given features $\mathcal{X}$ and labels $\mathcal{Y}$ is written as follows
\vspace{-.1cm}
\begin{align*}
\mathbb{P}\left[\mathcal{Y} | \mathcal{X}, T, \Theta \right]  
&= \prod_{\ell \in \mathcal{L}} \prod_{i \in \Delta(\ell)} \prod_{c \in \mathcal{C}} \theta_{\ell, (c)}^{\mathbb{I}^{(c)}(y_i)} \\
&= \prod_{\ell \in \mathcal{L}} \prod_{c \in \mathcal{C}} \theta_{\ell, c}^{\sum_{i \in \Delta(\ell)} \mathbb{I}^{(c)}(y_i)},
\end{align*}
where $\mathcal{L}$ is the set of leaves in the tree $T$, $\mathcal{C}$ is the set of classes, $\theta_{\ell, (c)}$ is the probability of sampling class $c$ under leaf $\ell$, and $\mathbb{I}^{(c)}(y_i)$ is the indicator function of whether $y_i$ belongs to class $c$. 
\end{propositionE}
\begin{proofE}
By construction, a decision tree $T$ partitions data into $|\mathcal{L}|$ splits, \ie, $\Delta(\ell)$ for $\ell \in \mathcal{L}$ where $\mathcal{L}$ is the set of leaves under the decision tree. Each leaf $\ell$ under the tree induces an independent probability distribution $\text{Dirichlet}(\boldsymbol{\theta}_\ell)$ over the probability of a given label $y_i$ occurring under leaf $\ell$. Along this line of reasoning, the probability of picking a given label $y_i$ is conditionally independent both of $\mathcal{X}$ and of the other classes in $\mathcal{Y}$ given the leaf $\ell$ it lands on (which is only determined from $T$ and $\mathbf{x}_i$) and the corresponding classification parameter $\boldsymbol{\theta}_\ell$ (which, again, is determined from $\Theta$ and $\ell$ only). The latter observation justifies all the steps in our proof, which we formulate in the following
\vspace{-.05cm}
\begin{align*}
    \mathbb{P}[\mathcal{Y}| \mathcal{X}, T, \Theta] 
    &= \prod_{i \in |\mathcal{Y}|} \mathbb{P}[y_i |\mathbf{x}_i, T, \Theta] \\ 
    &= \prod_{\ell \in \mathcal{L}} \prod_{i \in \Delta(\ell)} \mathbb{P}[y_i | \boldsymbol{\theta}_\ell] \\ 
    &= \boxed{\prod_{\ell \in \mathcal{L}} \prod_{c \in \mathcal{C}} \theta_{\ell, c}^{\sum_{i \in \Delta(\ell)} \mathbb{I}^{(c)}(y_i)}},
\end{align*}
which concludes the proof.
\end{proofE}

The Bayesian Classification and Regression Trees (BCART) construction \citep{chipman1998bayesian} assumes a Dirichlet prior distribution over $\Theta$, \ie, $\boldsymbol{\theta} \sim \text{Dirichlet}(\boldsymbol{\alpha})$ for all $\boldsymbol{\theta} \in \Theta$. Under this prior, we express the marginal likelihood $\mathbb{P}\left[\mathcal{Y} | \mathcal{X}, T \right]$ as follows.

\begin{propositionE}[Marginal Likelihood of a Bayesian DT][end,restate]\label{theorems::2} Assuming $\boldsymbol{\theta} \sim \text{Dirichlet}(\boldsymbol{\alpha})$, the marginal likelihood of a Bayesian decision tree $T$ given features $\mathcal{X}$ and labels $\mathcal{Y}$
\vspace{-.1cm}
\begin{align*}
\mathbb{P}\left[\mathcal{Y} | \mathcal{X}, T \right]  
&= \left(\frac{\Gamma(\sum_{c \in \mathcal{C}} \alpha_c)}{\prod_{c \in \mathcal{C}} \Gamma(\alpha_c)}\right)^{|\mathcal{L}|} \prod_{\ell \in \mathcal{L}} \frac{\prod_{c \in \mathcal{C}} \Gamma(n_{\ell, c} + \alpha_c)}{\Gamma(n_\ell + \sum_{c \in \mathcal{C}} \alpha_c)}, 
\end{align*}

where $n_{\ell, c} = \sum_{i \in \Delta(\ell)} \mathbb{I}^{(c)}(y_i)$ is the empirical count of data points with label $c$ at leaf $\ell$ and $n_{\ell} = \sum_{c \in \mathcal{C}} n_{\ell, c}$ is the total empirical count of data points (of all classes) at leaf $\ell$.
\end{propositionE}
\begin{proofE}
    We would like to derive $\mathbb{P}[\mathcal{Y}| \mathcal{X}, T]$ by marginalizing over $\Theta$. Assuming $\boldsymbol{\theta} \sim \text{Dirichlet}(\boldsymbol{\alpha})$, we prove the result of that, which we present in \cref{theorems::2}, in the following
\vspace{-.05cm}
\begin{align*}
    \mathbb{P}[\mathcal{Y}| \mathcal{X}, T] 
    &= \int_{\Theta} \mathbb{P}[\mathcal{Y}| \mathcal{X}, T, \Theta] \cdot p[\Theta] \: d\Theta\\ 
    &= \int_{\Theta} \prod_{\ell \in \mathcal{L}} \prod_{c \in \mathcal{C}} \theta_{\ell, c}^{\sum_{i \in \Delta(\ell)} \mathbb{I}^{(c)}(y_i)} \cdot \left(\frac{\theta_{\ell, c}^{\alpha_c}}{\frac{\prod_{c \in \mathcal{C}} \Gamma(\alpha_c)}{\Gamma(\sum_{c \in \mathcal{C}} \alpha_c)}} \right) \: d\Theta \\
    &= \left(\frac{\Gamma(\sum_{c \in \mathcal{C}} \alpha_c)}{\prod_{c \in \mathcal{C}} \Gamma(\alpha_c)}\right)^{|\mathcal{L}|} 
    \int_{\Theta} \prod_{\ell \in \mathcal{L}} \prod_{c \in \mathcal{C}} \theta_{\ell, c}^{\alpha_c + \sum_{i \in \Delta(\ell)} \mathbb{I}^{(c)}(y_i)} \: d\Theta \\ 
    &=  \left(\frac{\Gamma(\sum_{c \in \mathcal{C}} \alpha_c)}{\prod_{c \in \mathcal{C}} \Gamma(\alpha_c)}\right)^{|\mathcal{L}|} 
    \prod_{\ell \in \mathcal{L}} \int_{\boldsymbol{\theta}_\ell} \prod_{c \in \mathcal{C}} \theta_{\ell, c}^{\alpha_c + \sum_{i \in \Delta(\ell)} \mathbb{I}^{(c)}(y_i)} \: d\boldsymbol{\theta}_\ell \\ 
    &= \boxed{\left(\frac{\Gamma(\sum_{c \in \mathcal{C}} \alpha_c)}{\prod_{c \in \mathcal{C}} \Gamma(\alpha_c)}\right)^{|\mathcal{L}|} 
    \prod_{\ell \in \mathcal{L}} \frac{\prod_{c \in \mathcal{C}} \Gamma(n_{\ell, c} + \alpha_c)}{\Gamma(n_\ell + \sum_{c \in \mathcal{C}} \alpha_c)}}
\end{align*}
where $n_{\ell, c} = \sum_{i \in \Delta(\ell)} \mathbb{I}^{(c)}(y_i)$ is the empirical count of data points with label $c$ at leaf $\ell$ and $n_{\ell} = \sum_{c \in \mathcal{C}} n_{\ell, c}$ is the total empirical count of data points (of all classes) at leaf $\ell$, which concludes our proof.
\end{proofE}

We defer the proofs of the aforementioned results along with a more in-depth disussion of the role or priors to \cref{sec::app_likelihood}. As the prior over tree structures $\mathbb{P}\left[T | \mathcal{X} \right]$ remains a design choice, we defer discussing that to \cref{sec::mdp_reward}.

\textbf{Decision tree search space size.} 
The space of decision trees is enormous even for small depths and numbers of features. According to \citet{hu2019optimal}, assuming a (full) binary tree of depth $d_t$ and given a dataset with $p$ binary features, the number of distinct trees is 
\begin{align}
    N_{d_t} &= \sum_{n_0=1}^{1} \sum_{n_1=1}^{2n_0} \cdots \sum_{n_{d_t-1}=1}^{2n_{d_t-2}} 
    p \times \binom{2n_0}{n_1} (p-1)^{n_1} \times \cdots \nonumber \\
    &\quad \times \binom{2n_{d_t-2}}{n_{d_t-1}} (p - (d_t - 1))^{n_{d_t-1}}.
\end{align}
Hence, the size of the search space over decision trees up to some depth $d$ is simply $\sum_{d_t=1}^d N_{d_t}$. \cref{tab:search_space} computes the latter for different values of $d$ and $p$; exact computation for $d=5$ is already prohibitive, at least by elementary means.
\begin{table}[h!]
\centering
\begin{minipage}{0.44\linewidth}
\vspace*{1em}
\caption{Size of the search space over decision trees of depth $d \in \{1, 2, 3, 4, 5\}$ given a dataset with $p \in \{10, 20\}$ binary features.}
\label{tab:search_space}
\end{minipage}
\hfill
\begin{minipage}{0.54\linewidth}
\resizebox{1\linewidth}{!}{
\begin{tabular}{@{}cccc@{}}
\toprule
$d$ & \textbf{$p = 10$} & \textbf{$p = 20$} \\
\midrule
1 & $1.000 \times 10^{1}$ & $2.000 \times 10^{1}$ \\
2 & $1.000 \times 10^{3}$ & $8.000 \times 10^{3}$ \\
3 & $5.329 \times 10^{6}$ & $9.411 \times 10^{8}$ \\
4 & $5.609 \times 10^{13}$ & $8.358 \times 10^{18}$\\
\bottomrule
\end{tabular}
}
\end{minipage}
\end{table}

Given this challenge, we propose to use reinforcement learning methods to amortize search over this very large space, and the benefits of GFlowNets as amortized, diversity-seeking samplers.

\subsection{Amortized Inference with GFlowNets} \label{sec::gfn}

We briefly recall some of the main ideas behind GFlowNets relevant to our context, along with key ingredients to train them. A GFlowNet assumes access to a fully observable deterministic MDP with a set of states $\mathcal{S}$ and set of actions $\mathcal{A} \subseteq \mathcal{S} \times \mathcal{S}$. From the states in $\mathcal{S}$, we note in particular that the MDP $\mathcal{M}$ has a unique \textit{source state} $\boldsymbol{s}_0$ with no parents (represented as $T_0$ in \cref{fig::mdp_viz}), and a subset of states $\mathcal{X} \subset \mathcal{S}$ which we refer to as \textit{terminal states} (represented as $\perp$ in \cref{fig::mdp_viz}); terminal states have no outgoing actions and subsequently no children states. We assume that any state in $\mathcal{S}$ is reachable from $\boldsymbol{s}_0$ through some sequence of actions as illustrated in \cref{fig::mdp_viz}. We refer to a sequence of states $(\mathbf{s}_i \to \cdots \to \mathbf{s}_j)$ as a \textit{trajectory}, such that a transition between each pair of consecutive states $(\mathbf{s}_t \to \mathbf{s}_{t+1})$ is induced by an action $a \in \mathcal{A}$. In particular, we define a \textit{complete trajectory} as a trajectory that starts from $\boldsymbol{s}_0$ and ends at some $\boldsymbol{s}_n \in \mathcal{X}$, \ie, $\tau = (\mathbf{s}_0 \to \mathbf{s}_1 \to \cdots \to \mathbf{s}_n = \mathbf{x})$.

\paragraph{Policy.} A \textit{(forward) policy} operating on the aforementioned MDP outputs a distribution $\mathbb{P}_F[\mathbf{s'}| \mathbf{s}]$ for each state $\mathbf{s} \in \mathcal{S} \setminus \mathcal{X}$ over the states $\mathbf{s'}$ that are reachable from $\mathbf{s}$ within a single action\footnote{Note here that we can write the policy as a distribution over “next” states (from current state) or actions interchangeably as the MDP is deterministic.}. Given a complete trajectory $\tau$, a policy induces a distribution over $\tau$  written as 
\begin{equation}
\mathbb{P}_F\left[\tau = (\mathbf{s}_0 \to \mathbf{s}_1 \to \cdots \to \mathbf{s}_n = \mathbf{x})\right] = \prod_{t=0}^{n-1} \mathbb{P}_F\left[\mathbf{s}_{t+1}|\mathbf{s}_t\right].
\end{equation}
We denote the marginal distribution over terminal states by $\mathbb{P}_F^\top$. In general, estimating $\mathbb{P}_F^\top$ exactly or computing it in closed-form is intractable: 
\begin{equation}
\mathbb{P}_F^\top[\mathbf{x}] = \sum_{\tau \rightsquigarrow \mathbf{x}}{\mathbb{P}_F^\top\left[\tau\right]},
\end{equation}
where the sum is taken over all complete trajectories leading to $\mathbf{x}$.

Inheriting terminology from the RL literature, a \textit{reward function} $\mathcal{R}(\mathbf{x})$ for a GFlowNet represents the target---possibly and typically unnormalized---distribution over the set of terminal states $\mathcal{X}$. Formally, $\mathcal{R}$ is a mapping from the set of terminal states $\mathcal{X}$ to $ \mathbb{R}^+$, \ie, $\mathcal{R}: \mathcal{X} \to \mathbb{R}^+$. A typical form the reward function takes in our construction is $\mathcal{R}(\mathbf{x}) = e^{-\mathcal{E}(\mathbf{x})/T}$ where $\mathcal{E}: \mathcal{X} \to \mathbb{R}$ is an energy function and $T \in \mathbb{R}^+$ is a temperature control parameter. Given the latter, the learning problem a GFlowNet aims to approximate is sequentially choosing a policy $\mathbb{P}_F[\boldsymbol{s}_{t+1}| \boldsymbol{s}_t]$ such that the induced marginal distribution $\mathbb{P}_F^\top[\boldsymbol{s}_n = \boldsymbol{x}]$ is proportional to the reward function evaluated at $\mathbf{x}$ up to a normalizing constant, that is 
\begin{equation}\label{eq::gfn_reward}
    \mathbb{P}_F^\top[\boldsymbol{x}] \propto \mathcal{R}(\mathbf{x}) = e^{-\mathcal{E}(\mathbf{x})/T}.
\end{equation}
$\mathbb{P}_F$ is typically parametrized by a neural network with parameters $\boldsymbol{\theta}$, denoted $\mathbb{P}_F[\bullet\:;\: \boldsymbol{\theta}]$, that outputs the distribution over states $\mathbf{s'}$ reachable from any state $\mathbf{s} \in \mathcal{S}$ when taking argument $\mathbf{s'}| \mathbf{s}$. 

A key challenge within this line of framing is the potential intractability of the number of trajectories leading to some object $\mathbf{x}$, especially when the search space becomes combinatorially large and trajectories become longer. This makes $\mathbb{P}_F^\top$ and the normalizing constant in \eqref{eq::gfn_reward}, $Z = \sum_{\tau \rightsquigarrow \mathbf{x}} \mathcal{R}(\mathbf{x})$, hard to estimate. Various objectives have been proposed for circumventing this, usually via injecting additional learnable objects or parameters into the optimization problem. In particular, we use \textit{trajectory balance} \citep{malkin2022trajectory}, and introduce it in the following.

\paragraph{Trajectory balance (TB).} The TB objective is constructed by introducing an additional \textit{backward policy} $\mathbb{P}_B$, which can be learned or fixed, and a single scalar $Z_{\boldsymbol{\theta}}$ that estimates the partition function on the reward right hand side of \eqref{eq::gfn_reward}, \ie, $Z = \sum_{\tau \rightsquigarrow \mathbf{x}} \mathcal{R}(\mathbf{x})$. More precisely, $\mathbb{P}_B$ is defined as a collection of distributions $\mathbb{P}_B[\bullet| \mathbf{s}]$ over the parent states of any non-source state $\mathbf{s}$ and induces a distribution over complete trajectories $\tau$ leading to $\mathbf{x}$, \ie, 
\begin{equation}
\mathbb{P}_B\Big[\tau = (\mathbf{s}_0 \leftarrow \mathbf{s}_1 \leftarrow \cdots \leftarrow \mathbf{s}_n = \mathbf{x})| \mathbf{x}\Big] = \prod_{t=0}^{n-1} \mathbb{P}_B\Big[\mathbf{s}_{t}|\mathbf{s}_{t+1}\Big].
\end{equation}

Via this parametrization, the TB objective reduces enforcing \eqref{eq::gfn_reward} to (more simply) enforcing $\mathbb{P}_F[\tau] \propto \mathbb{P}_B[\tau| \mathbf{x}] \cdot \mathcal{R}(\mathbf{x})$ for every complete trajectory $\tau$ ending in $\mathbf{x}$ (note that the latter implies the former). Such proportionality is directly enforced by the training loss:
\begin{equation}\label{eq::tb}
    \ell_{\text{TB}}(\tau; \boldsymbol{\theta}) = \Big( \log\Big[Z_{\boldsymbol{\theta}}\cdot \mathbb{P}_F[\tau; \boldsymbol{\theta}]\Big] - \log\Big[\mathbb{P}_B[\tau| \mathbf{x}; \boldsymbol{\theta}] \cdot \mathcal{R}(\mathbf{x})\Big] \Big)^2.
\end{equation}

If $\ell_{\text{TB}}(\tau; \boldsymbol{\theta}) = 0$ for all complete trajectories $\tau$, then the output policy $\hat{\mathbb{P}}_F$ provably samples proportionally to the reward function, \ie, \eqref{eq::gfn_reward}, and the output normalization constant $Z_\theta$ equals the partition function of the reward, \ie, $Z_\theta = \sum_{\tau \rightsquigarrow \mathbf{x}} \mathcal{R}(\mathbf{x})$ \citep{malkin2022trajectory}. Another important resulting observation is that given some fixed $\mathbb{P}_B$, there exists a unique $\hat{\mathbb{P}}_F$ that satisfies \eqref{eq::gfn_reward}, which allows to set $\mathbb{P}_B$ to some fixed distribution at initialization.

\paragraph{Exploration in training.} The TB objective operates on trajectories in training, yet the process of choosing the trajectories to optimize \eqref{eq::tb} remains an algorithmic choice. A default choice is to train \textit{on-policy} by sampling $\tau \sim \mathbb{P}_F[\tau; \boldsymbol{\theta}]$ and minimizing $\ell_{\text{TB}}(\tau; \boldsymbol{\theta})$ with gradient descent. Given GFlowNets' aim to sample diversely from the reward function by design, favoring exploration in training has also proven to yield a variety of benefits. This could be done by sampling $\tau$ from a tempered $\mathbb{P}_F$ or through sampling individual actions from a uniform distribution $\epsilon$ of the time on average, akin to $\epsilon$-greedy exploration in RL. 

\section{Learning Decision Trees as Amortized Structure Inference}\label{sec::main}

\subsection{Constructing the GFlowNet's underlying MDP $\mathcal{M}$}

\textbf{State.} Given a choice of a maximum tree depth $d_{\text{max}}$ a tree is allowed to expand to, there are a maximum of $2^{d_{\text{max}}+1}-1$ nodes in $T$. We represent each node as a vector, which we order following a breadth-first traversal. Each node is either a decision node, in which case its corresponding vector is a decision rule, or a leaf node in which case it is labeled as an end-of-sentence (<EOS>). A decision rule is defined as a tuple $(f_i, t_i)$, where $f_i \in [d]$ is some given feature and $t_i \in \mathbb{R}$ is some real number corresponding to the splitting threshold for $f_i$. By notation, data-points satisfying $f_i \leq t_i$ flow to the left child node, and the rest flow to the right one. A choice at a child node is only allowed if a choice of a decision rule was made for all parent nodes. At initialization, each entry is assigned a “not yet specified” value of $\varnothing$. At the root node, either a decision rule or termination are chosen; if a decision rule is picked, $T$ is then recursively augmented in a similar way. 

\textbf{Action.} We opt to keep the action space discrete for simplicity. For that, we scale features back to $[0, 1]$ and pick $t$ thresholds uniformly spaced in this range; $t$ remains a hyperparameter of choice. An alternative paradigm could be using quantiles of training data, which we have found to often overfit the train set. As both the number of possible thresholds $f_i$ and $t_i$ can be large, we define the action space hierarchically. At each leaf node of each non-terminal tree state $T$, we can either pick a decision rule or terminate. Picking a decision rule consists in hierarchically choosing a feature $f_i$, then a threshold $t_i$. To account for decisions resulting in invalid states (see \cref{fig::mdp_viz}), we mask thresholds that do not result in a valid split, \ie, one where at least one data-point flows to each child node. A terminal state is reached either when an action to terminate is picked, or when all actions are masked.

\subsection{Reward Function and Parameter Sampling}\label{sec::mdp_reward}
\textbf{Reward.} As outlined in \cref{sec::setting}, we choose our reward function to be the joint conditional probability distribution
\begin{equation}
    \mathcal{R}(T | \mathcal{X}) =  \mathbb{P}\left[\mathcal{Y} | \mathcal{X}, T \right]  \cdot \mathbb{P}[T | \mathcal{X}] \propto \mathbb{P}[T| \mathcal{Y}, \mathcal{X}].
\end{equation}
\looseness=-1
Unlike a variety of settings where \textsc{GFlowNets} have been applied, the reward function can be written in closed form without any further parametrizations. The remaining ingredient is to choose a prior on the structure of $T$. Anchored in Occam's razor (\citet{rissanen1978modeling}; Chapter 28, \citet{mackay2003information}) and akin to previous work on (provably) optimal decision trees \citep{hu2019optimal, lin2020generalized, balcan2024learningaccurateinterpretabledecision}, we choose a prior of minimal (data) description length of $T$, as measured by its number of decision nodes. To properly normalize the likelihood term, we further inject a parameter $\beta$, and provide guidelines on tuning it appropriately. We formulate the posterior distribution over Bayesian DTs given our chosen prior $\mathbb{P}[T | \mathcal{X}] = e^{-\beta \cdot n(T)}$ as follows 
\begin{align*}
& \mathbb{P}[T |\mathcal{Y}, \mathcal{X}] \\
&\propto \mathbb{P}\left[\mathcal{Y} | \mathcal{X}, T \right]  \cdot \mathbb{P}[T | \mathcal{X}] \\
&\propto e^{-\beta \cdot n(T)} \cdot \left(\frac{\Gamma(\sum_{c \in \mathcal{C}} \alpha_c)}{\prod_{c \in \mathcal{C}} \Gamma(\alpha_c)}\right)^{|\mathcal{L}|} \prod_{\ell \in \mathcal{L}} \frac{\prod_{c \in \mathcal{C}} \Gamma(n_{\ell, c} + \alpha_c)}{\Gamma(n_\ell + \sum_{c \in \mathcal{C}} \alpha_c)} 
\end{align*}
where $\beta \geq 0$ and $n(T)$ is the number of decision (non-leaf) nodes in a tree $T$.

\textbf{Choice of $\beta$.} An important consideration when computing the prior over trees is the choice of the parameter $\beta$, which controls how much we penalize overly complex trees. For a choice of a large $\beta$, our model might not be expressive enough. For a choice of a small $\beta$, our model might become overly complex, potentially undermining its interpretability. We would like to pick some $\beta$ such that the resulting model would be expressive enough, have a short description length and be interpretable. Relying on information theoretical arguments, as outlined in \cref{pf::beta}, we propose that a suitable choice of $\beta$ is $\beta \sim \log(4) + \log(d) + \log(t)$, where $d$ is the number of splitting features, and $t$ is the number of splitting thresholds corresponding to the chosen discretization of the feature support. 

\textbf{Sampling classification parameters at inference.} Given a trained tree structure, classification parameters $\boldsymbol{\theta}_\ell$ at each leaf are sampled from the posterior $\boldsymbol{\theta}_\ell \sim \text{Dirichlet}(\boldsymbol{n}_\ell + \boldsymbol{\alpha})$, where $\boldsymbol{n}_\ell$ are the empirical counts of samples belonging to each class at leaf $\ell$ and $\boldsymbol{\alpha}$ is a prior on overall class counts.

\subsection{Parametrization of the Forward Policy}

A key property of our formulation is that it allows to frame decision tree learning as reasoning over root-to-leaf paths by picking a sequence of decision rules given a context of previous decisions. Given that both the prior and the predictions over different root-to-leaf paths are independent, a learned tree representation is invariant to their order.\footnote{This would make a sequence model, such as a (decoder-only) transformer \citep{vaswani2017attention, radford2018improving} or a order-invariant version \citep{lee2019set} a natural choice, but this choice turns out not to be computationally effective, especially at small-to-medium scales.} A representation given as input to our policy model is a set of (padded) vectors, each representing the root-to-leaf path to a leaf. The termination probability
$\mathbb{P}_F\big[\text{\small{<EOS>}}\big| \mathbf{s}_{t}\big]$ is parametrized as a simple multi-layer perceptron (MLP). The likelihoods of other actions
$\mathbb{P}_F\big[\mathbf{s}_{t+1}\big| \mathbf{s}_{t}; \neg \text{\small{<EOS>}} \big]$ are parametrized by a second MLP, evaluated independently on the representation of each leaf. The backward policy $\mathbb{P}_B$ is simply set to be uniform over parent states. To encourage exploration and diversely sample from the posterior, trajectories need not always be sampled from $\mathbb{P}_F$ (\emph{on-policy}). Instead, we use two widely adopted techniques from the RL literature, which have also been used for off-policy exploration in GFlowNets: a replay buffer and $\epsilon$-random exploration with annealed $\epsilon$. See 
 \cref{app::training_details} for all hyperparameters and training details.

\section{Empirical Evaluation}\label{sec::exps}
\begin{table*}[h!]
\begin{minipage}{\textwidth}
\centering
\caption{\textbf{Test accuracy and model size (total number of tree nodes) for single decision tree baselines, averaged over five random seeds}. For each algorithm, to account for model variance, we construct 1000 trees and pick the best tree in training. For Bayesian algorithms (including our \name), we choose the tree with the highest log-posterior. For all of the others, we choose the tree with the highest accuracy in training.}\vspace*{-1em}
\label{tab::benchmark_dt}
\begin{adjustbox}{max width=\textwidth}
\begin{tabular}{@{}l>{\centering\arraybackslash}p{1.8cm}>{\centering\arraybackslash}p{1.8cm}>{\centering\arraybackslash}p{1.8cm}>{\centering\arraybackslash}p{1.8cm}>{\centering\arraybackslash}p{1.8cm}>{\centering\arraybackslash}p{1.8cm}>{\centering\arraybackslash}p{1.8cm}>{\centering\arraybackslash}p{1.8cm}@{}}
\toprule
\textbf{Dataset} $\rightarrow$ & \multicolumn{2}{c}{\textbf{Iris}} & \multicolumn{2}{c}{\textbf{Wine}} & \multicolumn{2}{c}{\textbf{Breast Cancer}$^{\rm (D)}$} & \multicolumn{2}{c}{\textbf{Raisin}} \\
\cmidrule(lr){2-3} \cmidrule(lr){4-5} \cmidrule(lr){6-7} \cmidrule(lr){8-9}
\textbf{Algorithm} $\downarrow$ & \textsc{Test Acc}$\uparrow$ & \textsc{Size}$\downarrow$ & \textsc{Test Acc}$\uparrow$ & \textsc{Size}$\downarrow$ & \textsc{Test Acc}$\uparrow$ & \textsc{Size}$\downarrow$ & \textsc{Test Acc}$\uparrow$ & \textsc{Size}$\downarrow$  \\
\midrule
\textsc{Smc} & 0.9518 \std{0.02} & 16.18 \std{1.72}& 0.9311 \std{0.04} & 16.25 \std{2.66} & 0.931 \std{0.01} & 32.32 \std{2.68} & \cellcolor{blue!10}0.866 \std{0.01} & 46.58 \std{2.12} \\
\textsc{Mcmc} & 0.923 \std{0.04} & \cellcolor{blue!10}13.4 \std{1.5} & \cellcolor{blue!10}0.955 \std{0.02} & \cellcolor{blue!10}13.82 \std{1.21} & 0.92 \std{0.02} & 25.62 \std{2.67} & 0.864 \std{0.02} & 35.29 \std{1.93} \\
\textsc{Maptree} & 0.8733 \std{0.04} & 3.80 \std{0.45}& 0.9139 \std{0.02}& 4.8\std{0.45}& 0.9281 \std{0.02}& \cellcolor{blue!10}5 \std{0}& 0.8344 \std{0.03}& 7.80 \std{1.10}\\
\textsc{BCART} & 0.9267 \std{0.03} & 56.2 \std{31.8} &  0.9389 \std{0.02} & 49.8 \std{34.26} & 0.9018 \std{0.03} & 20.6 \std{12.09} & \cellcolor{blue!10}0.8678 \std{0.01} & 23.80 \std{8.26} \\
\midrule
\textsc{$(\alpha^*, \beta^*)$-Tsallis} & 0.9267 \std{0.04} & 16.6 \std{1.5} & 0.944 \std{0.315} & 19 \std{1.26} & \cellcolor{blue!10}0.937 \std{0.017} & 22.2 \std{0.98} & 0.864 \std{0.01} & 28.2 \std{1.6} \\
\textsc{CART-Gini} & 0.9494 \std{0.02} & 14.6 \std{1.5} &  0.876 \std{0.05} & 17.8 \std{4.12} & 0.923 \std{0.02} & 34.6 \std{6.25} & 0.852 \std{0.023} & 29.8 \std{0.98} \\
\textsc{CART-Entropy} & 0.9468 \std{0.02} & 14.6 \std{1.5} & 0.9357 \std{0.04} & 16.6 \std{3.2}& 0.9168\std{0.022}& 29.4 \std{1.96} & \cellcolor{blue!10}0.868 \std{0.016} & \cellcolor{blue!10}27.4 \std{0.08} \\
\midrule
\textsc{DPDT-4} & 0.947 \std{0.027} & 14.8 \std{0.98} & 0.889 \std{0.068} & 20.6 \std{2.94} & 0.919 \std{0.024} & 24.6 \std{2.33} &  0.853 \std{0.016} & 27.2 \std{2.03} \\
\midrule
\textsc{Quant-BnB} & \cellcolor{blue!10}0.953 \std{0.029} & 15 \std{0} & 0.817 \std{0.023} & 12.2 \std{6.26} & \cellcolor{blue!10}0.933 \std{0.023} & 15 \std{0} &  0.859 \std{0.011} & \cellcolor{blue!10}15 \std{0} \\
\midrule
\name (ours)  & \textcolor{lightblue}{\textbf{0.98}\std{0.04}} &  \cellcolor{blue!10}8.6\std{3.44} &  \textcolor{lightblue}{\textbf{0.97}\std{0.03}} &  \cellcolor{blue!10}8.6\std{1.5} & \textcolor{lightblue}{\textbf{0.95}\std{0.02}} & \cellcolor{blue!10}6.2\std{0.98} &  \textcolor{lightblue}{\textbf{0.9}\std{0.002}} & \cellcolor{blue!10}27\std{2.83} \\
\bottomrule
\end{tabular}
\end{adjustbox}
\end{minipage}

\vspace{0.5cm}

\begin{minipage}{\textwidth}
\centering
\caption{\textbf{Test accuracy and model size (total number of tree nodes; where applicable) for ensemble methods, averaged over five random seeds.} For tree methods, including ours, each ensemble contains 1000 trees.}\vspace*{-1em}
\label{tab::benchmark_ensemble}
\begin{adjustbox}{max width=\textwidth}
\begin{tabular}{@{}l>{\centering\arraybackslash}p{1.8cm}>{\centering\arraybackslash}p{1.8cm}>{\centering\arraybackslash}p{1.8cm}>{\centering\arraybackslash}p{1.8cm}>{\centering\arraybackslash}p{1.8cm}>{\centering\arraybackslash}p{1.8cm}>{\centering\arraybackslash}p{1.8cm}>{\centering\arraybackslash}p{1.8cm}@{}}
\toprule
\textbf{Dataset} $\rightarrow$ & \multicolumn{2}{c}{\textbf{Iris}} & \multicolumn{2}{c}{\textbf{Wine}} & \multicolumn{2}{c}{\textbf{Breast Cancer}$^{\rm (D)}$} & \multicolumn{2}{c}{\textbf{Raisin}} \\
\cmidrule(lr){2-3} \cmidrule(lr){4-5} \cmidrule(lr){6-7} \cmidrule(lr){8-9}
\textbf{Algorithm} $\downarrow$ & \textsc{Test Acc}$\uparrow$ & \textsc{Size}$\downarrow$ & \textsc{Test Acc}$\uparrow$ & \textsc{Size}$\downarrow$ & \textsc{Test Acc}$\uparrow$ & \textsc{Size}$\downarrow$ & \textsc{Test Acc}$\uparrow$ & \textsc{Size}$\downarrow$  \\
\midrule
\textsc{Greedy RF} & 0.96\std{0.01} & 14.73\std{0.9} & \cellcolor{blue!10}0.9833\std{0.01} & 18.89\std{0.78} & 0.9474\std{0.02} & 33.71\std{0.93} & 0.8689\std{0.01} & 151.12\std{2.53} \\
\textsc{XGBoost} & 0.9533\std{0.03} & 4.56\std{0.34} & 0.9556\std{0.03} & 3.51\std{0.08} & \cellcolor{blue!10}0.9579\std{0.02} & 6.86\std{0.16} & 0.8611\std{0.02} & 28.33\std{0.24} \\
\textsc{CatBoost} & 0.96\std{0.03} & 379.771\std{1.90} & 0.978\std{0.01} & 380.52\std{0.85}  & \cellcolor{blue!10}0.9544\std{0.01} & 126.72\std{0.4} & \cellcolor{blue!10}0.88\std{0.02} & 126.73\std{0.10} \\
\textsc{LightGBM} & 0.9533\std{0.03} & 9.39\std{0.58} & \textcolor{lightblue}{\textbf{0.9889\std{0.01}}} & 12.34\std{0.34} & \cellcolor{blue!10}0.9579\std{0.01} & 47.04\std{0.64} & 0.8622\std{0.02} & 59.63\std{0.17} \\
\midrule
\textsc{MLP} & \cellcolor{blue!10}0.9667\std{0.03} & N/A & 0.8500\std{0.15} & N/A & 0.9140\std{0.01} & N/A & 0.5167\std{0.03} & N/A \\

\textsc{TabTransformer} & 0.7933\std{0.049} & N/A & 0.978 \std{0.021} & N/A & 0.9316\std{.0211} & N/A & 0.8544 \std{0.03} & N/A \\
\textsc{FTTransformer} & 0.953\std{0.016} & N/A & 0.967\std{0.02} & N/A & \textcolor{lightblue}{\textbf{0.961}\std{0.013}} & N/A & 0.847\std{0.028} & N/A \\
\midrule
\name (ours)  & \textcolor{lightblue}{\textbf{0.973}\std{0.04}} &  \cellcolor{blue!10}11.11\std{2.94} &  \cellcolor{blue!10}0.983\std{0.01} &  \cellcolor{blue!10}8\std{1.29} & \cellcolor{blue!10}0.954\std{0.02} & \cellcolor{blue!10}5.67\std{0.19} &  \textcolor{lightblue}{\textbf{0.883}\std{0.03}} & \cellcolor{blue!10}26.31\std{1.25} \\
\bottomrule
\end{tabular}
\end{adjustbox}
\end{minipage}
\end{table*}
In \cref{sec::exp_dt}, we first evaluate our approach on two desirable properties when designing models for tabular data: 1) in-distribution generalization, as measured by held-out set accuracy, and 2) complexity (often also used as a proxy for interpretability), as measured by model size for DT-based methods, where model size directly correlates with model complexity. Next, in \cref{sec::exp_ensemble}, we would like to evaluate the potential of constructing ensembles of predictors sampled from \name, and how these can be competitive with gradient-boosted trees and deep learning methods. In \cref{sec::exp_sys_gen}, we consider experiments in systematic generalization, where we would like to see how models learned via structure inference with our method compare to a variety of other baselines. Finally, we conduct consistent scaling experiments in \cref{sec::exps_consistent_scaling}. We further show in \cref{app:budget-exps} that DT-GFN compares favorably to state-of-the-art at minimum cost (see \cref{fig::scaling}). For all experiments, we highlight the \textbf{\textcolor{lightblue}{best}} result in bold and blue and \colorbox{blue!10}{most competitive} (second best) result(s) in blue cell shades. For experiments in which we report model sizes, we omit mentioning the “best” result as it is hard to determine the optimal one. We further point to \cref{app:baseline_details} for details on baseline implementations where needed.

\subsection{Benchmarking with Single Decision Tree Algorithms}\label{sec::exp_dt}

\looseness=-1
\paragraph{Baselines and evaluation.} We compare our approach against 9 methods belonging to four different families: 1) Bayesian DT sampling algorithms: \textsc{Smc} \citep{lakshminarayanan2013topdownparticlefilteringbayesian}, \textsc{Mcmc} \citep{lakshminarayanan2013topdownparticlefilteringbayesian}, \textsc{MapTree} \citep{sullivan2023maptreebeatingoptimaldecision} and \textsc{BCART} \citep{chipman1998bayesian}; 2) methods with explicitly specified splitting criteria: \textsc{$(\alpha^*, \beta^*)$-Tsallis Entropy} \citep{balcan2024learningaccurateinterpretabledecision}, \textsc{CART-Gini}, \textsc{CART-Entropy}; 3) dynamic programming or RL-based methods: \textsc{Dpdt-4} \citep{kohler2024interpretabledecisiontreesearch}; 4) methods formulating decision tree learning as explicit optimization: \textsc{Quant-BnB} \citep{pmlr-v162-mazumder22a}. We conduct our series of experiments on a variety of widely used tabular datasets from the UCI repository \citep{dua2019uci}: Iris \citep{fisher1936use}, Wine \citep{aeberhard1992wine}, Breast Cancer Diagnostic \citep{dua2019uci} and Raisin \citep{guvenir2017raisin}. For each algorithm, we restrict the chosen maximum tree depth to 5 and we report the average held-out set accuracy and model size over five different train-test splits. Model size is measured by the total number of nodes in a tree. Results are presented in \cref{tab::benchmark_dt}.

\paragraph{Results.} In \cref{tab::benchmark_dt}, we observe that a decision tree sampled from a trained \name policy consistently—and often significantly—outperforms state-of-the-art single decision tree construction algorithms from all families. We also show that we obtain trees with the shortest data description length, measured by the average total number of nodes across seeds. Furthermore, when the maximum tree depth is sufficiently large, the minimal complexity comes from the ability of the model to abide by the prior rather than by explicit termination when reaching some small maximum depth.

\subsection{Benchmarking with Greedy Ensemble Methods and Deep Learning Methods}\label{sec::exp_ensemble}

\paragraph{Baselines.} We compare our approach to seven methods from two categories: (1) bootstrapped or gradient-boosted trees—\textsc{Greedy Rf} \citep{breiman2001random}, \textsc{XgBoost} \citep{chen2016xgboost}, \textsc{CatBoost} \citep{dorogush2018catboost}, and \textsc{LightGBM} \citep{ke2017lightgbm}; (2) deep learning models—\textsc{MLP}, \textsc{TabTransformer} \citep{huang2020tabtransformertabulardatamodeling}, and \textsc{FTTransformer} \citep{gorishniy2021revisiting}. The \textbf{experimental setup} and \textbf{evaluation criteria} follow \cref{sec::exp_dt}.

\paragraph{Results.} In \cref{tab::benchmark_ensemble}, we observe that an ensemble constructed from \name samples with Bayesian model averaging (as per \cref{alg:rf-gfn-prediction}) is consistently among the most competitive methods while keeping low model complexity. 

\subsection{Experiments in Systematic Generalization}\label{sec::exp_sys_gen}

\subsubsection{Robustness to Distribution Shifts}\label{sec::ood-gen} 

\looseness=-1
We consider distribution shifts along two features of the Pima Indians Diabetes dataset \citep{smith1988pima}: BMI and Age. The experimental setup is detailed in \cref{app::distribution_shift}. \textbf{Baselines} follow \cref{sec::exp_ensemble}. Ensembles of trees generated by \name compare favorably to baselines both in-distribution and out-of-distribution under both shifts (\cref{tab::ood_generalization}).

\begin{table}[!h]
\centering
\caption{\textbf{In-distribution and out-of-distribution test accuracies across two domain shifts.} For tree-based models, results for largest ensembles are reported. All ensembles contain 1000 trees.}\vspace*{-1em}
\begin{adjustbox}{max width=1\linewidth}
\begin{tabular}{@{}lcccc}
\toprule
\textbf{Domain shift feature} $\rightarrow$ & \multicolumn{2}{c}{\textbf{BMI}} & \multicolumn{2}{c}{\textbf{Age}} \\
\cmidrule(l){2-3} \cmidrule(l){4-5}
\textbf{Algorithm} $\downarrow$ & \textsc{In-dist.} & \textsc{OoD} & \textsc{In-dist.} & \textsc{OoD} \\
\midrule
\textsc{RF} & 0.803 & \cellcolor{blue!10}0.637 & \cellcolor{blue!10}0.8625 & \cellcolor{blue!10}0.66 \\
\textsc{XGBoost} & 0.77 & 0.598 & 0.788 & 0.61 \\
\textsc{CatBoost} & 0.803 & 0.606 & \cellcolor{blue!10}0.838 & \cellcolor{blue!10}0.645 \\
\textsc{LightGBM} & 0.803 & 0.59 & 0.825 & 0.624 \\
\midrule
\textsc{MLP} & 0.7541 & 0.57 & 0.825 & 0.513 \\
\textsc{TabTransformer} & 0.771 & 0.62 & 0.825 & 0.559 \\
\textsc{FTTransformer} & \cellcolor{blue!10}0.836 & 0.585 & 0.775 & 0.535 \\
\midrule
\name (ours) & \textcolor{lightblue}{\textbf{0.94}} & \textcolor{lightblue}{\textbf{0.755}} & \textcolor{lightblue}{\textbf{0.925}} & \textcolor{lightblue}{\textbf{0.7}} \\
\bottomrule
\end{tabular}
\end{adjustbox}
\label{tab::ood_generalization}
\end{table}

\begin{figure*}[t!]
    \centering
    \begin{minipage}{0.60\textwidth}
    \includegraphics[width=\textwidth,trim = 0 95 0 0,clip]{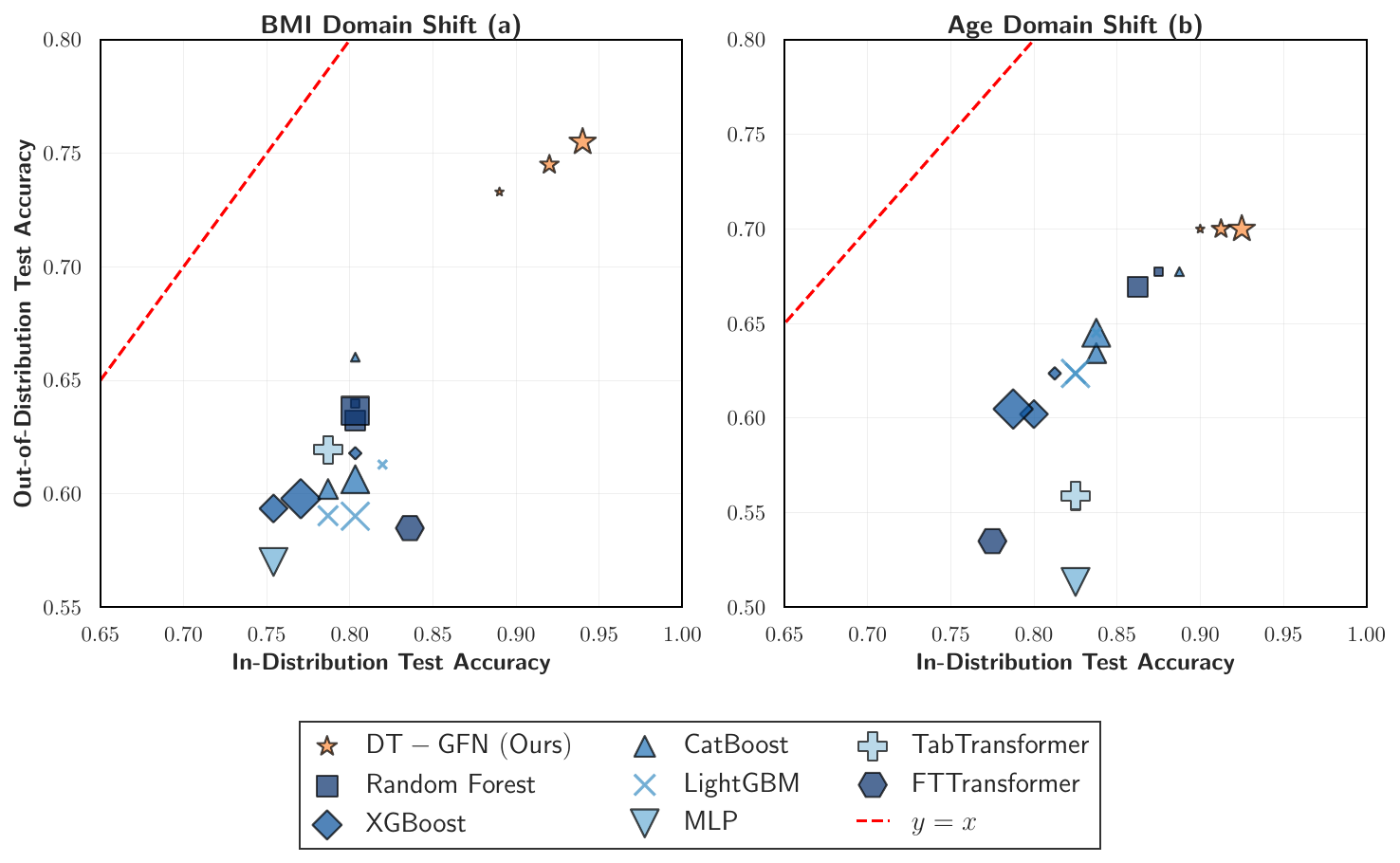}~\hspace*{2mm}~\raisebox{1.3in}{\includegraphics[width=0.63\textwidth,trim = 120 0 120 345,clip]{figures/scatter_subfigs.pdf}}
    \end{minipage}\hfill
    \begin{minipage}[t]{0.38\textwidth}
    \caption{\textbf{Distribution shift in-distribution/out-of-distribution plots with ablations on ensemble sizes $[100, 500, 1000]$ for tree-based methods.} Visualization of distribution shifts caused by interventions on (a) BMI features and (b) Age features, with the symbol sizes indicating the ensemble sizes.}
    \label{fig:dist_shifts}
    \end{minipage}
    \vspace*{-1em}
\end{figure*}

\subsubsection{Out-of-distribution Detection}\label{sec::ood-detection} 

\textbf{Task description.} An out-of-distribution detection task comprises a training set, only formed of the \emph{normal} class(es), and a test set that has a mix of normal and \emph{anomalous} classes; and consists in accurately predicting both.  

\textbf{Baselines.} We compare our approach against seven methods encompassing both recent deep learning algorithms and traditional approaches. Specifically, we include three deep learning methods: \citep{shenkar2022anomaly}, \textsc{Drocc} \citep{goyal2020drocc} and \textsc{Goad} \citep{bergman2020classification}, as well as five classical ML methods: \textsc{Copod} \citep{li2020copod}, \textsc{IForest} \citep{iforest}, \textsc{KNN} \citep{knn}, \textsc{PIDForest} \citep{gopalan2019pidforestanomalydetectionpartial}, and \textsc{RRCF} \citep{rrcf}. The \textbf{experimental setup} and all baseline results are drawn directly from \citet{shenkar2022anomaly}, as we are able to reproduce their exact results using the provided code.

\looseness=-1
\textbf{Evaluation.}  \citet{shenkar2022anomaly} assume \emph{a priori} knowledge of the number of anomalous samples in the test set, say $n_a$, following a common protocol in anomaly detection \citep{zong2018deep}. Then, they adjust the threshold on the prediction metric accordingly so as to select $n_a$ anomalous samples exactly. We outline the details of this procedure in \cref{app:exp_setup_details}. For evaluating our method, we alleviate the need for knowing $n_a$. For our experiments, as we have access to a trained \name policy that acts as a as probabilistic model, we can compute the probability of a sample to be normal/anomalous directly. This allows for setting-specific flexibility in designing a classifier, for instance whether we care more about overall accuracy or about being risk-averse, \ie, minimizing misdetections. In our case, we follow a simple procedure where we classify a sample as anomalous if its probability of being normal is at least two standard deviations lower than the average normal class classification probability across samples at test time. Following \citet{shenkar2022anomaly}, we use the F1 score as a metric.

\looseness=-1
\paragraph{Results.} In \cref{tab::ood_detection}, we observe that samples of single trees generated by \name perform (often significantly) better than the state-of-the-art in generalization to detect out-of-distribution samples, while still offering the benefits of a generative probabilistic model and without a priori knowledge of the number of anomalous samples in the test set.

\begin{table}[!h]
\centering
\caption{\textbf{F1 scores on common OOD detection benchmark datasets}, setting is reproduced as per the guidelines in \citet{shenkar2022anomaly}. \name results use a single tree.}\vspace*{-1em}
\resizebox{\linewidth}{!}{%
    \begin{tabular}{@{}lcccc}
    \toprule
    \textbf{Algorithm} $\downarrow$ \textbf{Dataset} $\rightarrow$ & \textbf{Thyroid} & \textbf{Ecoli} & \textbf{Vertebral} & \textbf{Glass} \\
    \midrule
    \textsc{Drocc} & 0.727 & N/A & 0.27 & \cellcolor{blue!10}0.222 \\
    \textsc{Goad} & 0.725 & 0.693 & \cellcolor{blue!10}0.269 & \cellcolor{blue!10}0.257 \\
    \citep{shenkar2022anomaly} & \cellcolor{blue!10}0.768 &\cellcolor{blue!10}0.7 & \cellcolor{blue!10}0.26 & \cellcolor{blue!10}0.272 \\
    \midrule
    \textsc{Copod} & 0.308 & 0.256 & 0.017 & 0.11 \\
    \textsc{IForest} & \cellcolor{blue!10}0.789 & 0.589 & 0.13 & 0.11 \\
    \textsc{KNN} & 0.573 & \cellcolor{blue!10}0.778 & 0.1 & 0.11 \\
    \textsc{PIDForest} & 0.72 & 0.256 & 0.12 & 0.089 \\
    \textsc{RRCF} & 0.319 & 0.289 & 0.08 & 0.156 \\
    \midrule
    \name (ours) & \textcolor{lightblue}{\textbf{0.794}} & \textcolor{lightblue}{\textbf{0.812}} & \textcolor{lightblue}{\textbf{0.404}} & \textcolor{lightblue}{\textbf{0.315}} \\
    \bottomrule
    \end{tabular}
}
\label{tab::ood_detection}\vspace*{-1em}
\end{table}
\subsection{Experiments in Consistent Scaling}\label{sec::exps_consistent_scaling}

\looseness=-1
We vary ensemble sizes in $\{100, 500, 1000\}$ for tree-based models in experiments in \cref{sec::ood-gen}. For each ensemble size, scaling is not only in the samples collected from \name at inference, instead we train a separate model for each ensemble size configuration. By ensemble size, we do not only mean the samples generated from the \name policy at inference to construct a predictive ensemble, instead we also describe by that the number of trees \name generates to compute the TB loss \eqref{eq::tb}. As TB is computed at the level of trajectories, computing it on more trees for the same number of training steps allows it to see more trajectories on average. During inference, we sample the same number of trees used in training. We observe in \cref{fig:dist_shifts} that ensembles constructed from trees generated by \name exhibit properties of systematic and consistent scaling in the ensemble size, \ie, more trees in an ensemble in training results in a increase in generalization both in-distribution and out-of-distribution across two distribution shift instances. On the other hand, gradient-boosted tree algorithms do not scale well with increasing the ensemble size, yielding unstable scaling behavior.

\section{Discussion and Future Work}

\name is an amortized inference method that generates decision tree models from the Bayesian posterior by sequential construction of decision rules. We have shown that \name is particularly strong in settings with few data points, scales well with problem and model size, and is effective in handling distribution shifts. Notable opportunities for future work include extending \name to online and streaming data scenarios \citep{pmlr-v238-chaouki24a}, where GFlowNets have been proposed as a solution \citep{dasilva2024streamingbayesgflownets}, and exploring its potential for knowledge-driven Bayesian model selection \citep{pmlr-v162-lotfi22a}.

\looseness=-1
The effectiveness of ensembles of decision trees in systematic generalization, as demonstrated by \name, makes a case for the use of amortized inference (\eg, using GFlowNets or other deep RL methods) to sample rule-based models: while a neural model is used for parameter inference and Bayesian model averaging, the generated classification models themselves remain lightweight and interpretable. In this sense, \name is a proof of concept for amortized inference of model structure in more general model classes. It should be built upon in future work on structure inference for probabilistic circuits (of which decision trees are a special case), probabilistic programs, and other structured models.


\if{0}
\begin{itemize}
    \item Very good at modeling from very few points or for datasets with effective feature dimension << actual feature dimension. 
    \item \textsc{TableShift}, modeling distribution shifts. 
    \item Tangible and systematic way to doing knowledge-driven/theoretical work on Bayesian model selection, for instance the type of work in \citep{pmlr-v162-lotfi22a}.
    \item Training generative tabular models on a mixture of datasets using an LLM feature encoder.
    \item Straightforward extension to the online setting, for instance relying on the setup and ideas of \citep{kohler2024interpretabledecisiontreesearch}.
    \item In settings with streaming data, so we can fine-tune on these \citep{dasilva2024streamingbayesgflownets}
\end{itemize}
\fi

\section*{Impact Statement}
This paper presents work whose goal is to advance the field of Machine Learning. There are many potential societal consequences of our work, none which we feel must be specifically highlighted here. 

\section*{Acknowledgments}

We thank Tristan Deleu for many valuable discussions in the course of this project. Y.B.\ acknowledges the support from CIFAR and the CIFAR AI Chair program as well as NSERC funding for the Herzberg Canada Gold medal. The research was enabled in part by computational resources provided by the Digital Research
Alliance of Canada (\url{https://alliancecan.ca}), Mila (\url{https://mila.quebec}), and
NVIDIA. 

\nocite{hernandez-garcia2024}
\bibliography{main_paper}
\bibliographystyle{icml2025}

\newpage
\appendix
\onecolumn
\addcontentsline{toc}{section}{Appendix} 
\color{black}\part{Appendix} 

\raggedbottom

\section{Illustrating Example- Pitfalls of Greedy Decision Trees and Ensemble Methods}\label{sec::hidden_xor}

To motivate part of our work, we devise a simple setting that showcases some of the pitfalls of greedy tabular methods, such as greedily constructed decision trees, and subsequent ensemble methods. We start by creating a variant of what is typically referred to as the hidden XOR problem, where we construct a synthetic dataset which has 20 features, 2 of which are binary and the rest are randomly generated real features. The label for each data point is simply an XOR operation between the two binary features, while the rest of the features are completely irrelevant for the classification task. Typically, a sufficiently expressive training dataset in this setting would have a minimum of 4 data points, enumerating all possible relevant input-output pairs, assuming we are not trying to test if a given model can generalize beyond the seen training distribution modes. In essence, we would like to examine a behavior trends that allow us to control the task difficulty, while observing how methods scale in that. In particular, we show how simply scaling the number of irrelevant features affects performance, with a dataset size of $|\mathcal{D}| = 1000$.

\begin{figure*}[h!]
    \centering
    \includegraphics[width=0.8\textwidth]{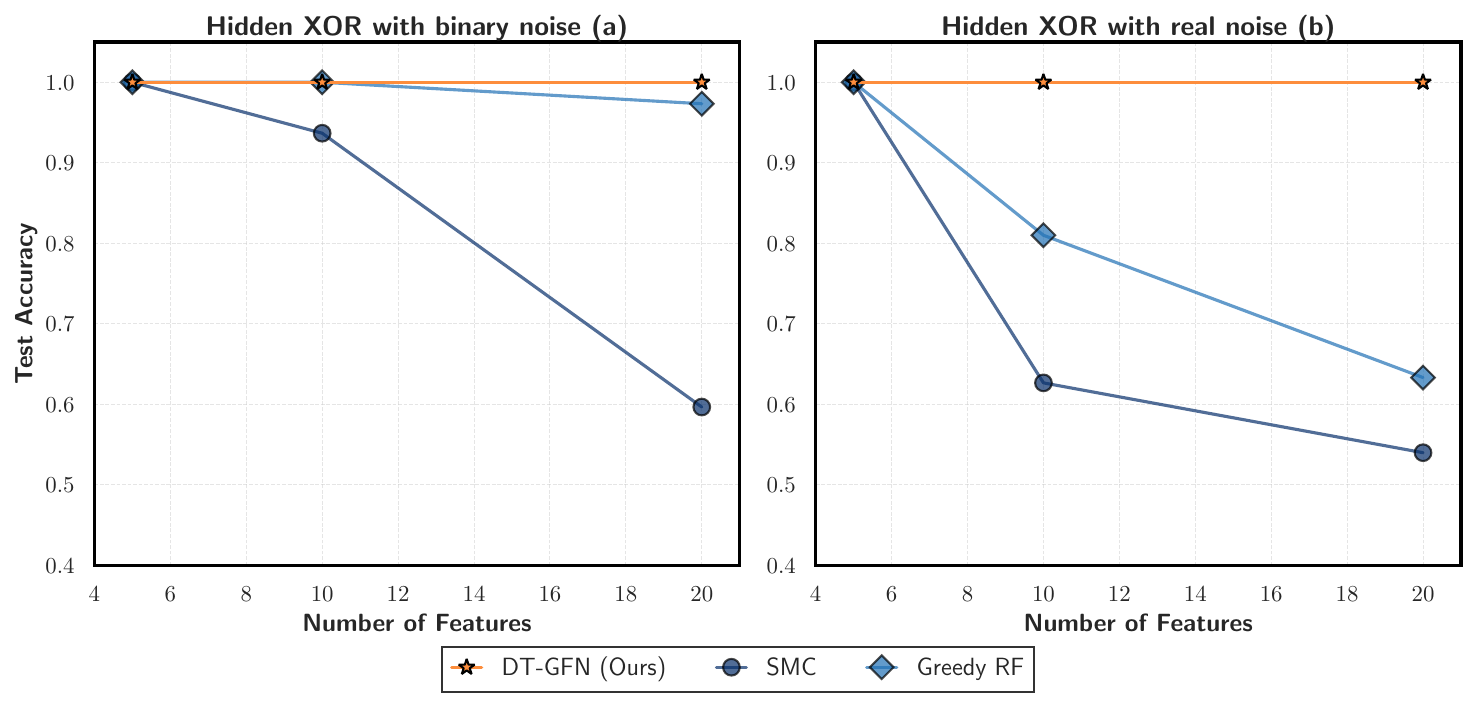}
    \caption{\textbf{Varying the number of features in a hidden XOR task where the label is an XOR operation between two features.} Noise features are chosen to be either binary (\textbf{left}) or real (\textbf{right}). All datasets contain 1000 samples.}
    \label{fig::xor}
\end{figure*}

\textbf{Scaling the number of noise features.} Varying the number of irrelevant or \textit{noise} features allows to progressively control the task difficulty, and examine how different methods behave under varying amounts of noise, \ie, a larger model search space, but no underlying structure being added to the data generating process. On the left of \cref{fig::xor}, the chosen noise features are binary, which significantly restricts the search space over decision trees. For instance, even when consider a total of 20 binary features, there are only 8,000 possible distinct decision trees of depth 2 that can be constructed given the data (a decision tree of depth 2 is enough to reconstruct the label generating process). We observe perfect test accuracy on all models for a small number of noise features. While a GFlowNets maintain perfect test accuracy as the number of noise features gets larger, we see a slight drop in test accuracy for greedy RFs for 20 features and even more significant drop for SMC as the number of features increases. Note that for 10 features, the number of possible distinct decision trees of depth 2 is only 1000 (as shown in \cref{tab:search_space}). On the right of \cref{fig::xor}, the noise features are now randomly generated real numbers,  which significantly enlarges the search space over decision tree models. We observe that test accuracy drops significantly for both greedy RFs and SMC when increasing the number of noise features, while GFlowNets maintain perfect test accuracy. We argue here, through presenting a simple task where the label generating process does not change but only the number of noise features increases, that structure inference is crucial to learning and mitigating many undesirable characteristics of greedy methods. 

\newpage

\newpage
\section{Mathematical Glossary}\label{sec::glossary}

We aim to make the manuscript rather self-contained, hence we introduce a few mathematical concepts that would provide a clear insight into some of the methods we consider throughout the paper, either as our own or ones that we benchmark with.

\textbf{$\Gamma$-function.} For a real number $z>0$, the $\Gamma$ function is defined as an improper integral as follows
$$
\Gamma(z)=\int_0^{\infty} t^{z-1} e^{-t} dt
$$

\textbf{Tsallis Entropy.} Let $P = \{p_i\}$ be a discrete probability distribution, where $p_i \geq 0$ and $\sum_i p_i = 1$. The Tsallis entropy of $P$ with entropic index $q \in \mathbb{R}$ is defined as
\begin{equation*}
S_q(P) = \frac{1}{q-1} \left( 1 - \sum_{i} p_i^q \right), \quad q \neq 1.
\end{equation*}
In the limit as $q \to 1$, Tsallis entropy reduces to the classical Shannon entropy
\begin{equation*}
S_1(P) = - \sum_{i} p_i \log p_i.
\end{equation*}

\section{Proofs and Auxiliary Results}\label{sec::theory}

We outline proofs of our theoretical claims, along with reasoning for some of the design choices that we use for our construction. 

\subsection{Proofs}\label{sec::app_likelihood}

We reiterate that the results in this section were already proved in \cite{chipman1998bayesian} in some form, we reproduce them given our setting and notation for the convenience of the reader.

\printProofs

\subsection{Prior over Decision Tree Structure and Choice of $\beta$}\label{pf::beta}

To choose an appropriate prior $\beta$, we need to ensure that $\beta$ properly normalizes the likelihood $\mathbb{P}[\mathcal{Y}| \mathcal{X}, T]$, \ie, accurately approximates the number of tree structures $T$ that would model $\mathcal{X}$. Alternatively, from a coding theory perspective, $\beta$ could also be interpreted as the average code length of a tree structure modeling $\mathcal{X}$. The number of possible binary tree structures with \( n \) internal nodes is given by the \( n \)-th Catalan number, \( C_n \), which asymptotically behaves as
   \[
   C_n \sim \frac{4^n}{n^{3/2}}.
   \]
   The corresponding coding length for that is
   \[
   \log(C_n) \approx n \log(4) - \frac{3}{2} \log(n).
   \]

Now, for each of the \( n \) nodes, we have to pick a splitting feature out of \(d \) variables and one of \( t \) possible thresholds, the complexity of such a choice on average is
   \[
   n (\log(p) + \log(t)).
   \]

Overall, the total coding length \( \ell(n) \) for a tree with \( n \) decision nodes is
\[
\ell(n) \approx n \Big(\log(4) + \log(p) + \log(t)\Big) - \frac{3}{2} \log(n).
\]

Hence a prior favoring trees with shorter description lengths can be expressed as 
\begin{align*}
\mathbb{P}[T| \mathcal{X}] & \propto \exp\Big(-\ell(n)\Big) \\
& \propto \exp\Big( -n \Big[\log(4) + \log(p) + \log(t)\Big] + \frac{3}{2} \log(n) \Big)
\end{align*}

Considering the asymptotically dominant term, we should choose $\beta$ such that
\begin{align*}
\boxed{\beta \sim \log(4) + \log(p) + \log(t)}
\end{align*}

\subsection{Reward Computation in Mini-Batches}

Computing the reward for a given tree, as per \cref{sec::mdp_reward}, requires recursing over the tree $|\mathcal{X}|$ times in the worst case. Given how large $|\mathcal{X}|$ in our considered setting can get, an important result to scaling our approach is the ability to compute rewards in mini-batches. As shown already in \citep{bengio2023gflownet}, computing the rewards in mini-batches ensures that the optimal policy $\hat{\mathbb{P}}_F$ minimizing the TB square-loss for all complete trajectories converges to the true reward/posterior in expectation.

\section{Construction of Ensembles of Predictors}
Given access to a trained \name policy, we perform a prediction using samples from that as follows. 

\begin{algorithm}[h!]
\caption{Bayesian Ensemble Prediction with \name Samples}
\label{alg:rf-gfn-prediction}
\begin{algorithmic}
\State \textbf{Input.} Data point $\mathbf{x}_j$, set of decision tree samples $\{T_i\}$, dataset $\mathcal{D}$ 
\State \textbf{Output.} Predicted class $\hat{y}_j$

\For{each tree $T_i$}
    \State Compute $\mathbb{P}(y_j = c \mid \mathbf{x}_j, T_i)$ for all classes $c$ using the GFlowNet policy.
\EndFor

\State Compute $\log\left[\mathbb{P}(T_i | \mathcal{D})\right]$ for all trees $T_i$
\State $m \gets \max\limits_{i} \log\left[\mathbb{P}(T_i | \mathcal{D})\right]$
\State $\log\left[\mathbb{P}(\mathcal{D})\right] \gets \log\left(\sum\limits_{k} \exp(\log\left[\mathbb{P}(T_k | \mathcal{D})\right] - m)\right) + m$

\For{each tree $T_i$}
    \State $\mathbb{P}(T_i | \mathcal{D}) \gets \frac{\exp\left(\log\left[\mathbb{P}(T_i | \mathcal{D})\right]\right)}{\sum\limits_{k} \exp\left(\log\left[\mathbb{P}(T_k | \mathcal{D})\right]\right)}$
\EndFor

\For{each class $c$}
    \State $\mathbb{P}(y_j = c | \mathbf{x}_j, \mathcal{D}) \gets \sum\limits_{i} \mathbb{P}(T_i | \mathcal{D}) \cdot \mathbb{P}(y_j = c | \mathbf{x}_j, T_i)$
\EndFor

\State $\hat{y}_j \gets \arg\max\limits_c \mathbb{P}(y_j = c | \mathbf{x}_j, \mathcal{D})$
\State \Return $\hat{y}_j$
\end{algorithmic}
\end{algorithm}

\newpage
\section{Training Details}\label{app::training_details}

We list a variety of strategies for GFlowNet training we use throughout the paper along with setup-specific hyperparameters allowing to reproduce our results. 

\subsection{Exploration Strategy}

To allow for exploration throughout training, we employ the following strategies, which we find to consistently perform well across our considered datasets and tasks. 

\textbf{$\epsilon$-greedy annealing.} We generate a set of trajectories from the \name policy throughout training, such that actions are sampled with probability $1-\epsilon\:$ according to the policy and with probability $\epsilon$ uniformly at random. $\epsilon$ is varied throughout training from some $\epsilon_{0}$ predefined at initialization to some small positive constant, \ie, $\epsilon \in (0, \epsilon_0]$. $\epsilon_0$ is set to 0.1 in our experiments.

\textbf{Replay buffer.} We use a replay buffer to store the Top-K “best" trees (with highest rewards) we have seen so far in training. Then, we sample trajectories from the latter using the backward policy $P_B$, simply set to a uniform distribution over parent states at each step starting from a given terminal state. 

\subsection{Hyperparameters}

We list hyperparameters we consistently use throughout our experiments in \cref{tab::hyperparams}. We highlight that it is also possible to use smaller stopping depths for faster training, or a smaller discretization threshold constant for datasets lower precisions, for instance only 1 for binary hidden XOR or 9 for Iris.

\begin{table}[h]
\centering
\caption{\textbf{Training hyperparameters for reproducing our experiments.}}\vspace*{-1em}
\begin{adjustbox}{max width=.5\textwidth}
\begin{tabular}{ll}
\toprule
\textbf{Hyperparameter} & \textbf{Value} \\
\midrule
\textbf{Tree Construction} \\
Max Tree Depth & 5 \\
Thresholds Discretization & 99 \\
Number of Samples & 1000 \\
\midrule
\textbf{Policy} \\
Policy Model & MLP \\
Hidden Layers & 3 \\
Hidden Units per Layer & 256 \\
\midrule
\textbf{Optimization} \\
Learning Rate & 0.01 \\
Training Steps & 100 \\
Batch Size (Forward) & 90 \\
Batch Size (Backward Replay) & 10 \\
\midrule
\textbf{Exploration} \\
Replay Buffer Capacity & 100 \\
Random Action Probability & 0.1 \\
\midrule
\textbf{Proxy} \\
Parameter Prior $(\boldsymbol{\alpha})$ & $\boldsymbol{\alpha} = [0.1]\times$ (number of classes)  \\
Structure Prior $(\beta)$ & $\log(4) + \log(d)$ \\
\bottomrule
\end{tabular}
\label{tab::hyperparams}
\end{adjustbox}
\end{table}

\newpage 

\section{Empirical Cost Analysis}

\subsection{Budget-constrained Training}\label{app:budget-exps}

We would like to test how DT-GFN would perform under a tight time/compute budget. As inference costs are negligible compared to training costs (see \cref{tab:cost_ensemble_sizes}), it would be interesting to observe: (i) how the accuracy of our predictions vary with training time budgets of $[0, 10, 20]$ (seconds) (ii) how ensemble predictions with \cref{alg:rf-gfn-prediction} vary in ensemble size given varying levels of structure inference training, from no time (random structure from a base DT-GFN policy) to a DT-GFN policy trained for 20 seconds. Experiment are carried on the Iris dataset, in the same setup as \cref{sec::exp_ensemble}; results are averaged over data split seeds $[1,2,3,4,5]$. 

We observe that DT-GFN manages to get high test accuracies, with \textbf{consistent scaling across both training time and ensemble size}. Notably, even with random structure, ensembles constructed according to \cref{alg:rf-gfn-prediction} still manage to scale consistently. For the maximum training budget of 20 seconds, all DT-GFN ensembles outperform the best gradient-boosted tree baseline (GBT) and match the best baseline in \cref{sec::exp_ensemble}. We further highlight that at a budget of 50 seconds, test accuracy plateaus at the current maximum for all ensemble sizes. Yet, these results are still lower than DT-GFN's results in \cref{sec::exp_ensemble} with more resources (see \cref{tab::hyperparams}), further hinting at scaling capacity with increase in resources.

\begin{figure*}[h!]
    \centering
    \includegraphics[width=0.6\textwidth]{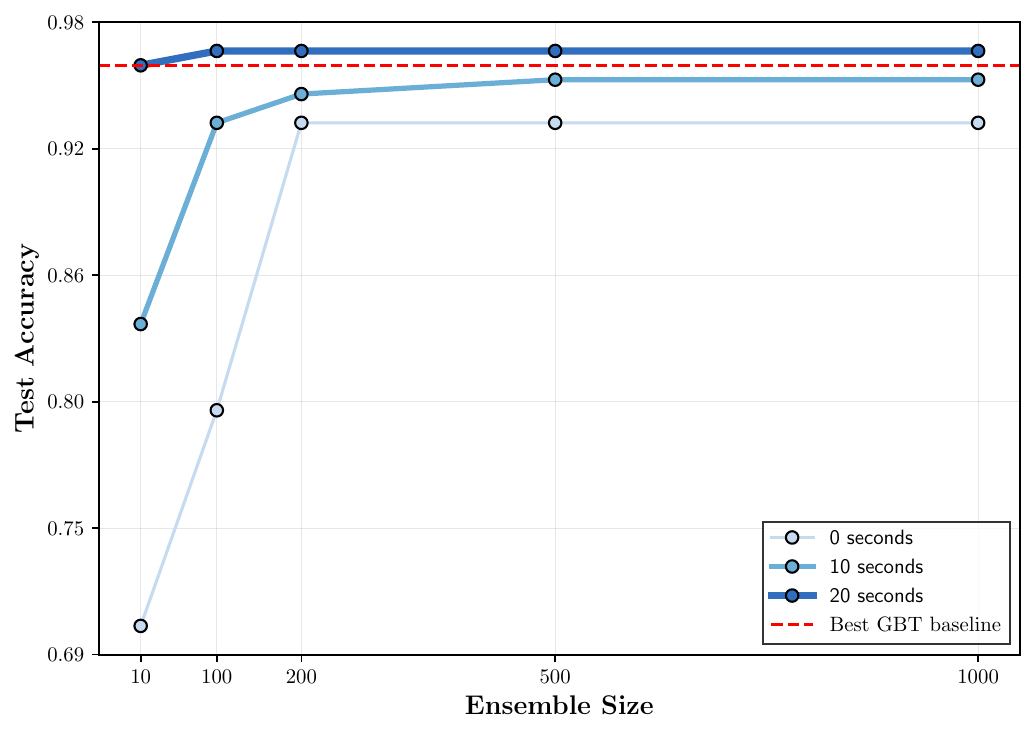}
    \caption{\textbf{DT-GFN scaling with ensemble size in $[10, 100, 200, 500, 1000]$ and allocated time/compute budget in [0 seconds, 10 seconds, 20 seconds]}. Experiment performed on the Iris dataset and results are averaged over data split seeds $[1,2,3,4,5]$. We show \textbf{consistent} scaling across both ensemble sizes and training time/compute resources. Inference costs are negligible as shown in \cref{tab:cost_ensemble_sizes}.}
    \label{fig::scaling}
\end{figure*}

\paragraph{Reproducibility.} Hyperparameters to reproduce our results are similar to \cref{tab::hyperparams}, with exceptions highlighted in \cref{tab:cost_hyperparams}. 

\begin{table}[h]
\centering
\caption{\textbf{Training hyperparameters for reproducing our experiments.}  \$\{\texttt{variable}\} denotes the varying scaling axes of \cref{fig::scaling}}\vspace*{-1em}
\begin{tabular}{ll}
\toprule
\textbf{Hyperparameter} & \textbf{Value} \\
\midrule
\textbf{Tree Construction} \\
Max Tree Depth & 3 \\
Thresholds Discretization & 9 \\
Number of Samples & \$\{\texttt{variable}\} \\
\midrule
\textbf{Optimization} \\
Training Steps & \$\{\texttt{variable}\} \\
\midrule
\textbf{Exploration} \\
Replay Buffer Capacity & 10 \\
Random Action Probability & 0.05 \\
\bottomrule
\end{tabular}
\label{tab:cost_hyperparams}
\end{table}

\subsection{Finite-time Amortization Cost Measurements}\label{app::cost}

We highlight how training and inference costs scale in the most “costly” training parameters we observe, which are tree max. depths and the number of trees (proportionally trajectories) the DT-GFN policy is trained on.

\begin{table}[h!]
\centering
\caption{\textbf{Training cost per epoch and inference cost (in seconds) for sampling an ensemble of DTs of varying size at different maximum depths.} }\vspace*{-1em}
\label{tab:cost_ensemble_sizes}
\begin{adjustbox}{max width=1\textwidth}
\begin{tabular}{@{}l>{\centering\arraybackslash}p{1.8cm}*{4}{>{\centering\arraybackslash}p{1.5cm}}@{}}
\toprule
& & \multicolumn{4}{c}{\textbf{Ensemble Size}} \\
\cmidrule(lr){3-6}
& & \multicolumn{2}{c}{\textsc{1}} & \multicolumn{2}{c}{\textsc{100}} \\
\cmidrule(lr){3-4} \cmidrule(lr){5-6}
\textbf{Dataset} & \textbf{Max. Depth} & \textbf{Train} & \textbf{Infer.} & \textbf{Train} & \textbf{Infer.} \\
\midrule
\multirow{2}{*}{Iris} & \textsc{2} & 1.09 & 0.0074 & 1.09 & 0.4239 \\
 & \textsc{4} & 4.21 & 0.0169 & 4.29 & 0.4468 \\
\midrule
\multirow{2}{*}{Wine} & \textsc{2} & 1.36 & 0.0032 & 1.85 & 0.283 \\
 & \textsc{4} & 3.95 & 0.0047 & 4.50 & 0.4953 \\
\midrule
\multirow{2}{*}{$\text{Breast Cancer}^{\rm (D)}$} & \textsc{2} & 1.57 & 0.0073 & 2.26 & 0.195 \\
 & \textsc{4} & 3.31 & 0.0145 & 3.73 & 0.5303 \\
\midrule
\multirow{2}{*}{Raisin} & \textsc{2} & 1.79 & 0.015 & 2.47 & 0.23 \\
 & \textsc{4} & 5.05 & 0.0037 & 6.23 & 0.4804 \\
\bottomrule
\end{tabular}
\end{adjustbox}
\end{table}

\subsection{Hardware}
Our training is conducted on an RTX 8000 GPU with 16GB of allocated memory. The training cost is measured on this hardware setup. Notably, the training process is lightweight and can be easily fitted on a local machine with minimal computational resources.

\section{Further Details on Experimental Setup}

\subsection{Datasets}\label{app::datasets}

Each dataset contains both numerical and categorical features, and the task in all datasets is classification. For the feature support to be tractable for discretization, we scale all features to $[0, 1]$ with Min-Max scaling and discretization is performed uniformly.   We list the datasets and splits we use in \cref{tab::dataset-aux}. 

\begin{table}[h!]
    \centering
    \small 
    \caption{\textbf{Dataset characteristics and reproducibility.} For systematic generalization experiments, test columns in the form $a/b$ denote number of $(a)$ in-distribution and $(b)$ out-of-distribution samples respectively.}
    \label{tab:dataset-comparison}
    \setlength{\tabcolsep}{6pt}  
    \begin{tabular}{@{}lcccccl@{}}  
        \hline
        \textbf{Dataset} & \textbf{n} & \textbf{d} & \textbf{Train} & \textbf{Test} & \textbf{Split Seeds} & \textbf{Experiment} \\
        \hline
        Iris \citep{fisher1936use} & 150 & 4 & 120 & 30 & [1, 2, 3, 4, 5] & DT \& Ensemble \\
        Wine \citep{aeberhard1992wine} & 178 & 13 & 142 & 36 & [1, 2, 3, 4, 5] & DT \& Ensemble\\
        $\text{Breast Cancer}^{\rm (D)}$  \citep{wolberg1995breast} & 569 & 30 & 455 & 114 & [1, 2, 3, 4, 5] & DT \& Ensemble\\
        Raisin \citep{guvenir2017raisin} & 900 & 7 & 720 & 180 & [1, 2, 3, 4, 5] & DT \& Ensemble \\
        \midrule
        Pima \citep{smith1988pima} \textbf{(BMI)} & 768 & 8 & 243 & 60/465 & [42] & Domain shift \\
        Pima \citep{smith1988pima} \textbf{(Age)} & 768 & 8 & 316 & 80/372 & [42] & Domain shift \\
        \midrule
        Thyroid \citep{quinlan1987thyroid} & 3772 & 6 & 1840 & 1839/93 & \cite{shenkar2022anomaly} & OOD detection  \\
        Ecoli \citep{horton1996probabilistic} & 336 & 7 & 164 & 163/9 & \cite{shenkar2022anomaly} & OOD detection\\
        Vertebral \citep{vertebral} & 240 & 6 & 106 & 104/30 & \cite{shenkar2022anomaly} & OOD detection \\
        Glass \citep{evett1987glass} & 214 & 9 & 103 & 102/9 & \cite{shenkar2022anomaly} & OOD detection \\
        \bottomrule
    \end{tabular}
\end{table}

\subsection{Preprocessing}\label{app::prep}
For the first two sets of datasets, we use \textsc{MinMax} scaling for scaling features back to $[0, 1]$. All categorical features are encoded using \textsc{OneHotEncoding}. For the third set of datasets, which we use for OOD detection, we rely on the preprocessing protocol of \citep{shenkar2022anomaly}. All datasets are obtained from the UCI repository \citep{dua2019uci} except the data for ODD detection we take it directly from \citep{shenkar2022anomaly}

\paragraph{Distribution shift procedure :} \label{app::distribution_shift}
We consider two distribution shifts in the Pima Indians Diabetes dataset \citep{smith1988pima}: (1) BMI Shift, where the training set includes patients with BMI < 30, and evaluation is performed on held-out sets with BMI < 30 and BMI > 30; and (2) Age Shift, where the training set consists of patients younger than 29 (median), with evaluation on held-out sets of age $\leq$ 29 and age > 29.

\paragraph{Out-of-distribution Detection Procedure :} Datasets are partitioned into training and test sets, where the training set consists exclusively of normal samples, while the test set includes both normal and anomalous samples. 
To account for variance and improve generalization, we leverage the \textit{bagging effect} by averaging scores obtained from multiple feature permutations. While this approach proves particularly beneficial for small \( d \) and very small \( n \), it comes at the cost of increased computational overhead. 
\citep{shenkar2022anomaly} generate multiple feature permutations, with the number of permutations, denoted \( P \), given by:

\[
P = \min\left(\left\lfloor \frac{100}{\log(n) + d} \right\rfloor + 1, 2\right)
\]

where \( n \) represents the number of training samples and \( d \) denotes the number of features.

\section{Auxiliary Experiments} \label{app:exp_setup_details}

\subsection{Additional Bayesian Decision Tree Baselines}\label{add::bayesian_exps}

As it was hard to reproduce some of the algorithms below, namely ones in \citet{cochrane2024divideconquercombinebayesian}, we test ours against them in the setting of \citet{cochrane2024divideconquercombinebayesian}. We elaborate on that below. The experimental setup and \name details are identical to those in \cref{sec::exp_dt}, unless otherwise stated in \cref{tab::dataset-aux}

\begin{table}[h!]
    \centering
    \caption{\textbf{Dataset characteristics.}}\vspace*{-1em}
    \label{tab::dataset-aux}
    \begin{tabular}{lcccccl}
        \hline
        \textbf{Dataset} & \textbf{n} & \textbf{d} & \textbf{Train} & \textbf{Test} & \textbf{Split Seeds} & \textbf{Experiment} \\
        \hline
        Iris \citep{fisher1936use} & 150 & 4 & 105 & 45 & [123456789] & DT \\
        Wine \citep{aeberhard1992wine} & 178 & 13 & 124 & 54 & [123456789] & DT \\
        $\text{Breast Cancer}^{\rm (O)}$ \citep{wolberg1995breast} & 699 & 9 & 489 & 210 & [123456789] & DT \\
        Raisin \citep{guvenir2017raisin} & 900 & 7 & 630 & 270 & [123456789] & DT\\
        \bottomrule
    \end{tabular}
\end{table}

In \cref{tab::results-aux}, using the same experimental setup as \citet{cochrane2024divideconquercombinebayesian}, we observe that samples from a single tree generated by \name perform comparably—or often significantly better—than state-of-the-art generalization baselines.

\begin{table}[H]
\centering
\caption{\textbf{Benchmarking} \name
\textbf{with Bayesian decision tree baselines in} \citet{cochrane2024divideconquercombinebayesian}, in the same setting as the latter.\label{tab::results-aux}}\vspace*{-1em}
\begin{adjustbox}{max width=.8\textwidth}
\begin{tabular}{@{}lcccc@{}}
\toprule
\textbf{Algorithm} $\downarrow$ \textbf{Dataset} $\rightarrow$ & \textbf{Iris} & \textbf{Wine} & \textbf{Breast Cancer}$^{\rm (O)}$ & \textbf{Raisin} \\
\midrule
\textsc{BCART} \citep{chipman1998bayesian} & 
\begin{tabular}[c]{@{}c@{}} 
0.908\std{0.022}
\end{tabular} & 
\begin{tabular}[c]{@{}c@{}} 
0.916\std{0.046}
\end{tabular} & 
\begin{tabular}[c]{@{}c@{}} 
0.939\std{0.014}
\end{tabular} & 
\begin{tabular}[c]{@{}c@{}} 
0.843\std{0.010}
\end{tabular} \\
\textsc{Smc} \citep{lakshminarayanan2013topdownparticlefilteringbayesian} & 
\begin{tabular}[c]{@{}c@{}} 
0.909\std{0.022}
\end{tabular} & 
\begin{tabular}[c]{@{}c@{}} 
0.978\std{0.022}
\end{tabular} & 
\begin{tabular}[c]{@{}c@{}} 
0.924\std{0.010}
\end{tabular} & 
\begin{tabular}[c]{@{}c@{}} 
0.842\std{0.010}
\end{tabular} \\
\textsc{Wu} \citep{wu2007bayesian} & 
\begin{tabular}[c]{@{}c@{}} 
N/A
\end{tabular} & 
\begin{tabular}[c]{@{}c@{}} 
N/A
\end{tabular} & 
\begin{tabular}[c]{@{}c@{}} 
0.922\std{0.017}
\end{tabular} & 
\begin{tabular}[c]{@{}c@{}} 
0.843\std{0.012}
\end{tabular} \\
\textsc{Hmc-Df} \citep{cochrane2023rjhmctreeexplorationbayesiandecision} & 
\begin{tabular}[c]{@{}c@{}} 
0.906\std{0.026}
\end{tabular} & 
\begin{tabular}[c]{@{}c@{}} 
0.950\std{0.039}
\end{tabular} & 
\begin{tabular}[c]{@{}c@{}} 
0.940\std{0.010}
\end{tabular} & 
\begin{tabular}[c]{@{}c@{}} 
0.847\std{0.004}
\end{tabular} \\
\textsc{Hmc-Dfi} \citep{cochrane2023rjhmctreeexplorationbayesiandecision} & 
\begin{tabular}[c]{@{}c@{}} 
0.917\std{0.023}
\end{tabular} & 
\begin{tabular}[c]{@{}c@{}} 
0.948\std{0.022}
\end{tabular} & 
\begin{tabular}[c]{@{}c@{}} 
0.952\std{0.007}
\end{tabular} & 
\begin{tabular}[c]{@{}c@{}} 
0.838\std{0.007}
\end{tabular} \\
\textsc{Dcc-Tree} \citep{cochrane2024divideconquercombinebayesian} & 
\begin{tabular}[c]{@{}c@{}} 
0.911\std{1.2e-16}
\end{tabular} & 
\begin{tabular}[c]{@{}c@{}} 
0.958\std{0.02}
\end{tabular} & 
\begin{tabular}[c]{@{}c@{}} 
0.952\std{0.004}
\end{tabular} & 
\begin{tabular}[c]{@{}c@{}} 
0.844\std{0.002}
\end{tabular} \\
\midrule
\name (ours) & 
\begin{tabular}[c]{@{}c@{}} 
\textcolor{lightblue}{\textbf{0.977}} 
\end{tabular} & 
\begin{tabular}[c]{@{}c@{}} 
\textcolor{lightblue}{\textbf{0.981}} 
\end{tabular} & 
\begin{tabular}[c]{@{}c@{}} 
\textcolor{lightblue}{\textbf{0.98}} 
\end{tabular} & 
\begin{tabular}[c]{@{}c@{}} 
\textcolor{lightblue}{\textbf{0.856}} 
\end{tabular} \\
\bottomrule
\end{tabular}
\end{adjustbox}
\label{tab::benchmark_bayesian}
\end{table}

\subsection{In-distribution/Out-of-distribution Plots with Ensemble Size Ablations} 

\begin{figure*}[h!]
    \centering
    \includegraphics[width=\textwidth]{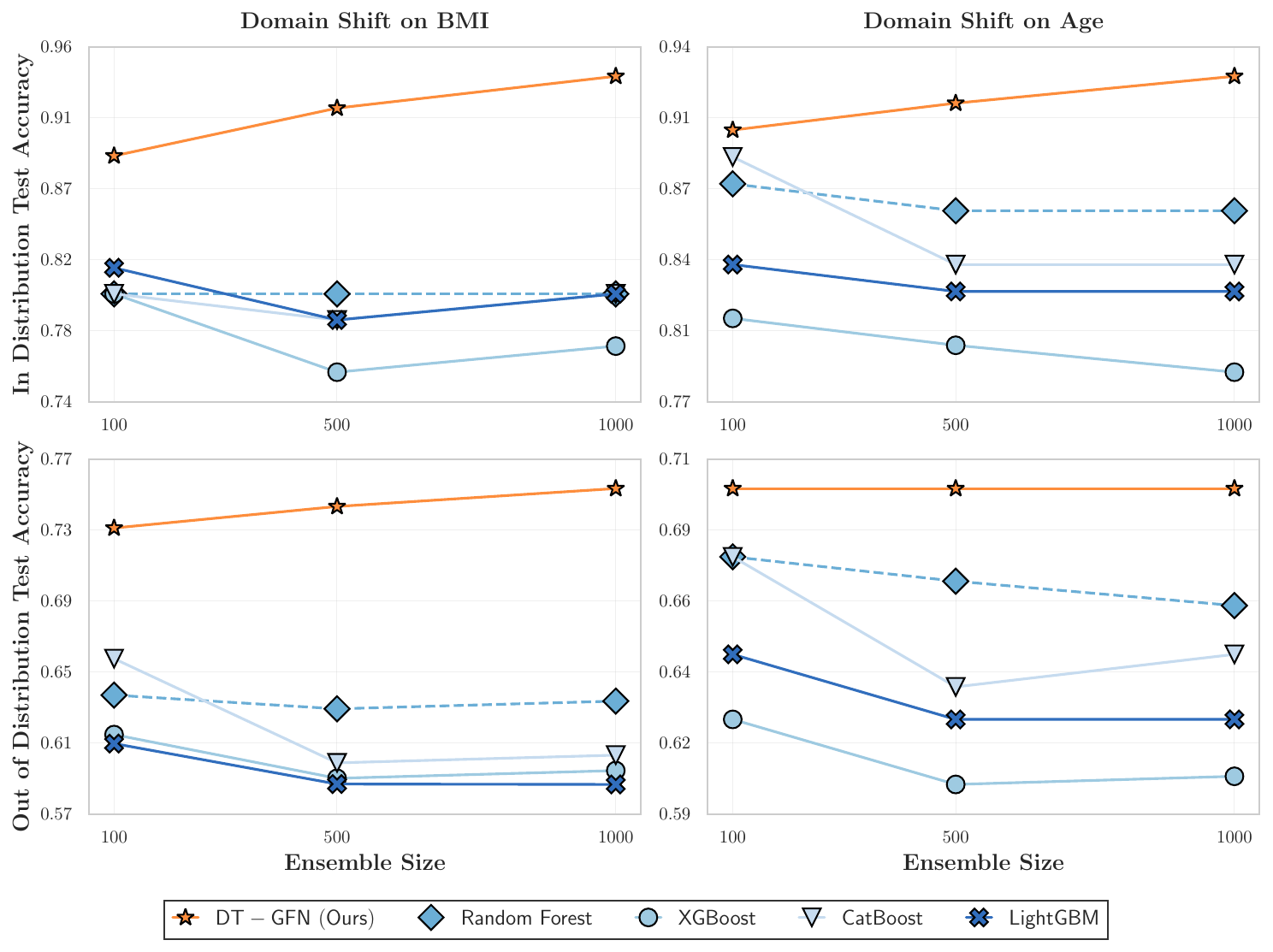}
        \caption{\textbf{Systematic increase in generalization accuracy both in-distribution and out-of-distribution in the ensemble size for tree-based methods.}}
    \label{fig::ablation_ensemble_size}
\end{figure*}

We more clearly visualize ablations in ensemble size for tree-based methods from \cref{fig:dist_shifts} in \cref{fig::ablation_ensemble_size}.

\newpage

\section{Baseline Details}\label{app:baseline_details}

We list important intuitions and/or reproducibility guidelines for some of the baselines we compare with, and further provide code for baselines in our code base at \href{https://github.com/GFNOrg/dt-gfn}{https://github.com/GFNOrg/dt-gfn}.

\textbf{Sequential Monte Carlo (SMC) Trees \citep{lakshminarayanan2013topdownparticlefilteringbayesian}.} SMC Trees is a Bayesian decision tree method that employs a Sequential Monte Carlo (SMC) approach to optimize tree construction. The method maintains a population of particles, each representing a candidate split defined by a feature and threshold, and iteratively updates them based on a Dirichlet likelihood and a prior on tree complexity. The model's hyperparameters—$\alpha$, which controls the Dirichlet prior, and $n_{\text{particles}}$, which regulates exploration—determine the trade-off between search diversity and convergence speed. The maximum depth is set to 5.

\textbf{MAPTree \citep{sullivan2023maptreebeatingoptimaldecision}.} MAPTree is a Bayesian decision tree method that constructs decision trees by performing maximum a posteriori (MAP) inference over a posterior distribution of tree structures and parameters. This approach balances model complexity and predictive performance by optimizing the posterior distribution of both tree structures and parameters. As mentioned in the original paper, we impose a 300-second time limit and systematically vary the number of tree expansions (\eg, 10, 100, 1000, 10000) to explore the trade-off between search depth and runtime. Additionally, we select $(\alpha, \beta, \rho)$ values to influence the prior distribution and regularization strength in the Bayesian framework.

\textbf{$(\alpha^*, \beta^*)$-Tsallis Entropy \citep{balcan2024learningaccurateinterpretabledecision}.} Tsallis entropy, as introduced in \cref{sec::glossary}, is a generalization of Shannon entropy used in information theory, parametrized by $\alpha^*$ and $\beta^*$, which control the degree of non-extensivity. In this context, the Tsallis entropy-based decision tree method introduces these entropy measures into the tree construction process, influencing the selection of splits. The method optimizes for both diversity and accuracy in the node partitions, making it particularly suited for data with complex or hierarchical structures. The model's hyperparameters ($\alpha^*$, $\beta^*$) are selected to balance between exploration of the feature space and exploitation of high-accuracy splits.

\textbf{Best $(\alpha^*, \beta^*)$ Search \citep{balcan2024learningaccurateinterpretabledecision}.} We follow the grid search guidelines in \citet{balcan2024learningaccurateinterpretabledecision} to find the best $(\alpha^*, \beta^*)$ combination given training data. In particular, we use the following $(\alpha^*, \beta^*)$ tuples for each of our datasets. 

\begin{table}[h!]
\centering
\begin{tabular}{l*{5}{c}} 
\toprule
\textbf{Dataset} $\downarrow$ \textbf{Seed} $\rightarrow$       & \textbf{1} & \textbf{2} & \textbf{3} & \textbf{4} & \textbf{5} \\
\midrule
Iris   &      $(0.5, 1)$      &     $(0.5, 1)$       &      $(0.5, 2)$      &      $(0.5, 4)$      &      $(2, 1)$      \\
Wine   &        $(1.1, 7)$ & $(0.5, 1)$ &    $(0.5, 1)$      &   $(0.5, 1)$   &      $(0.5, 1)$     \\
Breast Cancer Diagnostic  & $(2, 1)$ &   $(2, 2)$ &  $(7, 1)$   & $(2, 2)$ &  $(6, 1)$   \\
Raisin & $(2, 2)$ &   $(2, 1)$ &  $(2, 1)$   & $(2, 1)$ &  $(4, 1)$  \\
\bottomrule
\end{tabular}
\caption{\textbf{Best} $(\alpha^*, \beta^*)$ \citep{balcan2024learningaccurateinterpretabledecision} \textbf{for each considered dataset and split seed in \cref{sec::exp_dt}.}}
\label{tab:example}
\end{table}

\textbf{DPDT-4 \citep{kohler2024interpretabledecisiontreesearch}.} DPDT-4 is a parameterized decision tree algorithm that introduces a flexible partitioning strategy to create deeper and more expressive trees. The "4" in DPDT-4 refers to the maximum depth of the trees, which is designed to balance between model complexity and interpretability. This method focuses on minimizing the depth of the tree while maintaining high predictive accuracy, making it computationally efficient for large datasets.


\end{document}